\documentclass[10pt,twocolumn,letterpaper]{article}
\usepackage{iccv}
\usepackage{times}
\usepackage{epsfig}
\usepackage{graphicx}
\usepackage{amsmath}
\usepackage{amssymb}

\usepackage{multirow}
\usepackage{multicol}
\usepackage{booktabs}
\usepackage{subfigure}
\usepackage{microtype}
\usepackage{algorithm}  
\usepackage{algpseudocode}  
\usepackage{engord}

\usepackage[pagebackref=true,breaklinks=true,letterpaper=true,colorlinks,bookmarks=false]{hyperref}

\iccvfinalcopy 

\def\iccvPaperID{xxxx} 

\ificcvfinal\fi

\begin{document}

\title{TS-CAM: Token Semantic Coupled Attention Map for Weakly Supervised\\ Object Localization}

\author{Wei Gao$^1$\quad Fang Wan$^1$\quad Xingjia Pan$^{2,3}$\quad Zhiliang Peng$^1$\quad Qi Tian$^4$\quad \\ Zhenjun Han$^1$\quad Bolei Zhou$^5$\quad Qixiang Ye$^1$\\
$^1$University of Chinese Academy of Sciences\quad $^2$Youtu Lab, Tencent\quad \\$^3$NLPR, Institute of Automation, CAS\quad  $^4$Huawei Cloud AI \quad \\$^5$The Chinese University of Hong Kong\\

{\tt\small \{vasgaowei, xjia.pan\}@gmail.com}, {\tt\small \{wanfang, hanzhj, qxye\}@ucas.ac.cn}\\
{\tt\small tian.qi1@huawei.com}, {\tt\small bzhou@ie.cuhk.edu.hk}

}

\maketitle
\ificcvfinal\thispagestyle{empty}\fi

\begin{abstract}

   Weakly supervised object localization (WSOL) is a challenging problem when given image category labels but requires to learn object localization models. 
   Optimizing a convolutional neural network (CNN) for classification tends to activate local discriminative regions while ignoring complete object extent, causing the partial activation issue. 
   In this paper, we argue that partial activation is caused by the intrinsic characteristics of CNN, where the convolution operations produce local receptive fields and experience difficulty to capture long-range feature dependency among pixels. 
   We introduce the token semantic coupled attention map (TS-CAM) to take full advantage of the self-attention mechanism in visual transformer for long-range dependency extraction.
   %
   TS-CAM first splits an image into a sequence of patch tokens for spatial embedding, which produce attention maps of long-range visual dependency to avoid partial activation.
   TS-CAM then re-allocates category-related semantics for patch tokens, enabling each of them to be aware of object categories.
   TS-CAM finally couples the patch tokens with the semantic-agnostic attention map to achieve semantic-aware localization.
   Experiments on the ILSVRC/CUB-200-2011 datasets show that TS-CAM outperforms its CNN-CAM counterparts by $7.1\%/27.1\%$ for WSOL, achieving state-of-the-art performance. Code is available at \href{https://github.com/vasgaowei/TS-CAM}{https://github.com/vasgaowei/TS-CAM}
   %
   %
   
\end{abstract}

\section{Introduction}
\label{sec:intro}
Weakly supervised learning refers to methods that utilize training data with incomplete annotations to learn recognition models. Weakly supervised object localization (WSOL) solely requires the image-level annotations indicating the presence or absence of a class of objects in images to learn localization models~\cite{papandreou2015weakly,pathak2015constrained,roy2017combining,xue2019danet}. WSOL has attracted increasing attention as it can leverage the rich Web images with tags to learn object-level models~\cite{xue2019danet}.  

\begin{figure}[t]
    \centering
    \includegraphics[width=1.\linewidth]{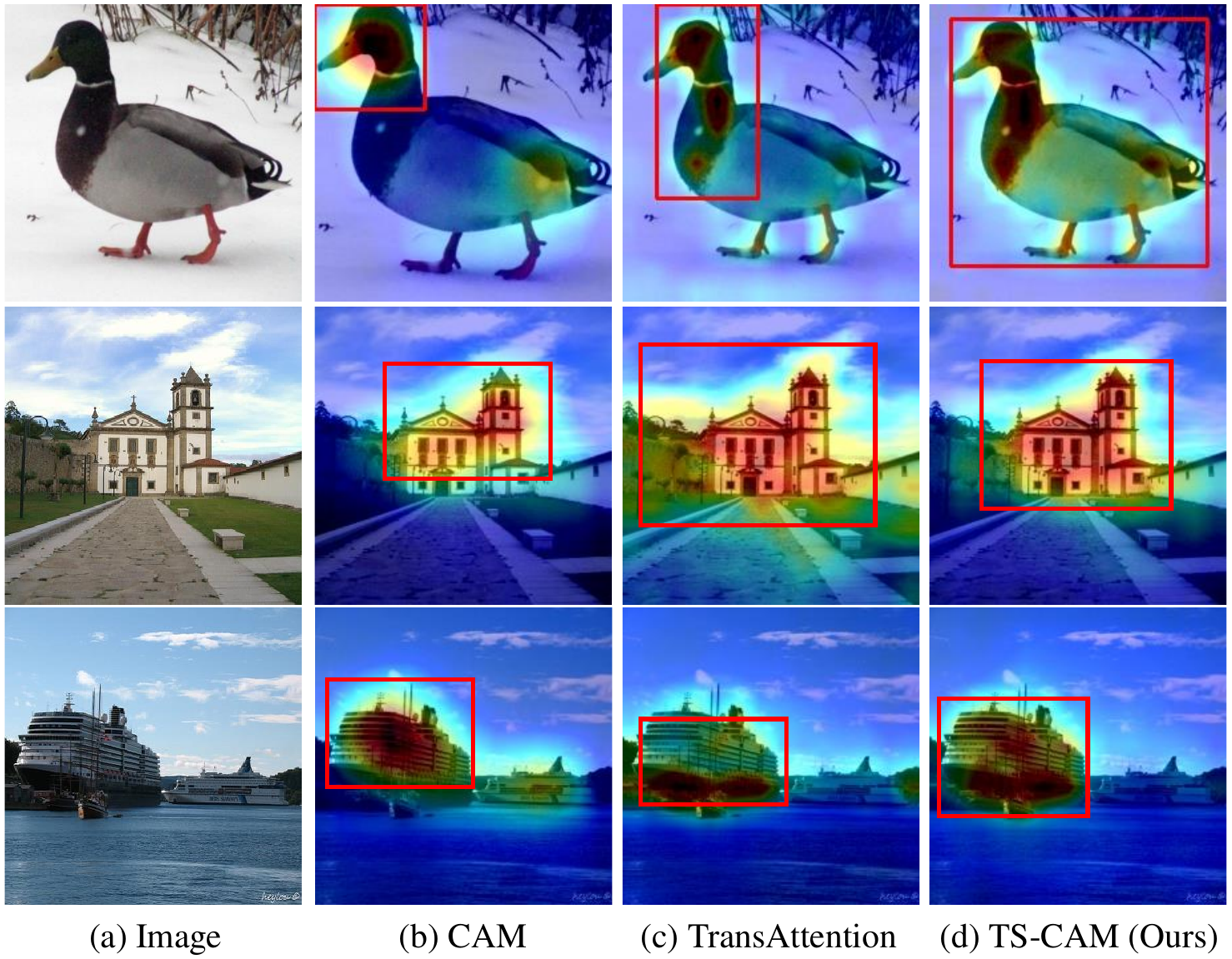}
    \caption{Comparison of weakly supervised object localization results. (a) Input Image. (b) Class Activation Map (CAM). (c) TransAttention: Transformer-based Attention. (d) TS-CAM. Object localization boxes are in red. (Best viewed in color)}
    \label{fig:motivation}
\end{figure}

\begin{figure*}[t]
    \centering
    \includegraphics[width=1.\linewidth]{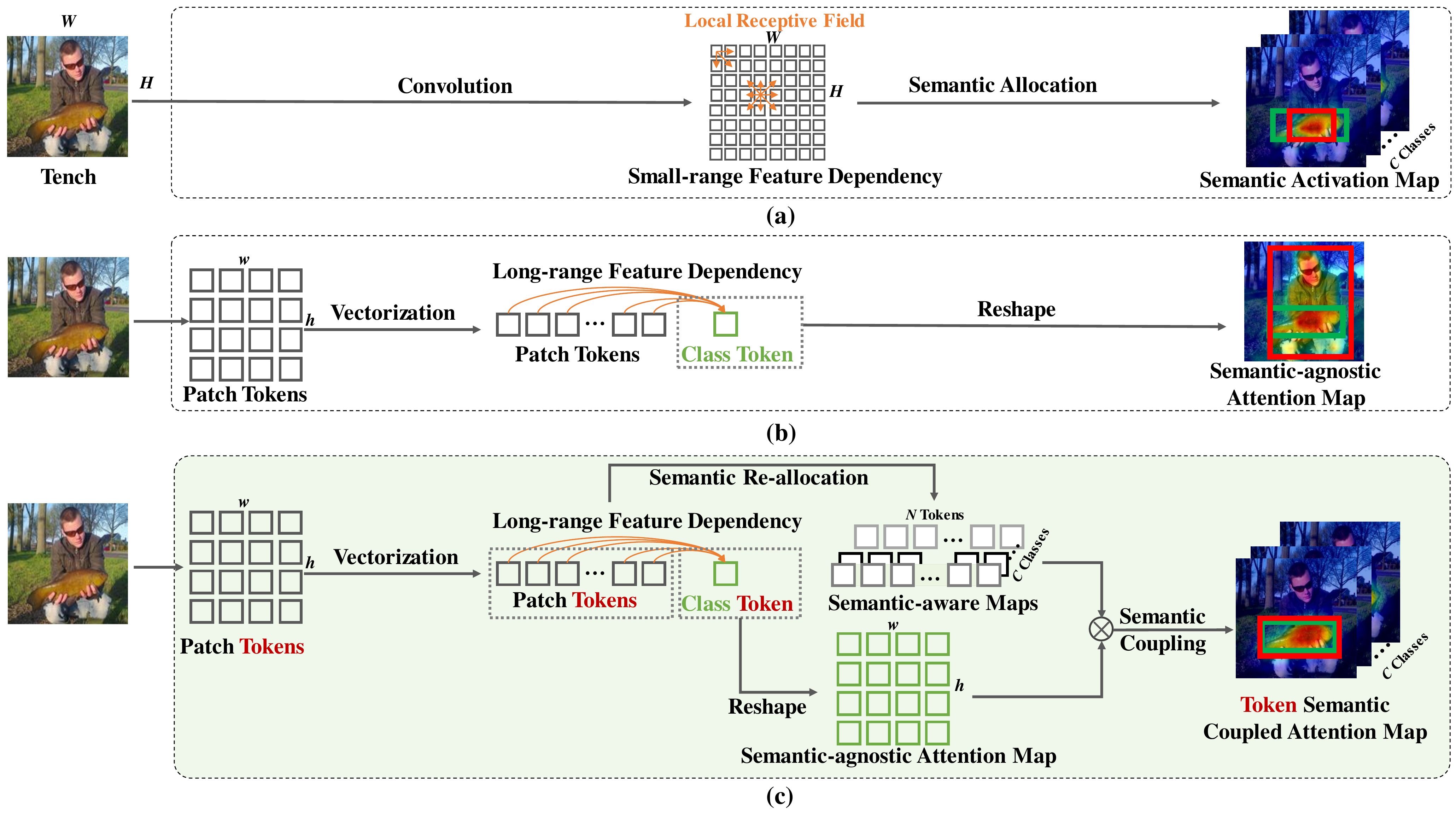}
    \caption{Comparison of the mechanisms of (a) CNN-based CAM, (b) Transformer-based Attention and (c) the proposed TS-CAM. The CNN-based CAM method is limited by the small-range feature dependency and the transformer-based attention is limited by the semantic-anostic issue. TS-CAM is able to produce semantic coupled attention maps for complete object localization. (Best viewed in color) }
    \label{fig:Flowchart_comparison}
\end{figure*}

As the cornerstone of WSOL~\cite{choe2020evaluating}, the Class Activation Mapping (CAM)~\cite{zhou2016learning} utilizes the activation map from the last convolution layer to generate semantic-aware localization maps for object bounding-box estimation.
However, CAM suffers from severe underestimation of object regions because the discriminative regions activated through the classification models are often much smaller than objects’ actual extent~\cite{Rethinking20}. Local discriminative regions are capable of minimizing image classification loss, but experience difficulty for accurate object localization~~\cite{xue2019danet}, Fig.~\ref{fig:motivation}(b).
Much effort has been made to solve this problem by proposing various 
regularizations~\cite{zhang2018adversarial,zhang2018self,xue2019danet,lu2020geometry,mai2020erasing}, divergent activation~\cite{singh2017hide,xue2019danet,yun2019cutmix} or adversarial training~\cite{choe2019attention,mai2020erasing,wei2017object,zhang2018adversarial,yun2019cutmix,singh2017hide}.
However, there is very little work to pay attention to fundamentally solving the inherent defects of CNN's local representation, Fig.~\ref{fig:Flowchart_comparison}(a). Capturing the long-range feature dependency, which can be interpreted as the semantic correlation between features in different spatial locations, is critical for WSOL.

Recently, visual transformer has been introduced to the computer vision area. Visual transformer~\cite{vit2020} constructs a sequence of tokens by splitting an input image into patches with positional embedding and applying cascaded transformer blocks to extract visual representation. Thanks to the self-attention mechanism and Multilayer Perceptron (MLP) structure, visual transformers can learn complex spatial transforms and reflect long-range semantic correlations adaptively, which is crucial for localizing full object extent, Fig.~\ref{fig:motivation}(d).
Nevertheless, visual transformer cannot be directly mitigated to WSOL for the following two reasons: (1) When using patch embeddings, the spatial topology of the input image is destroyed, which hinders the generation of activation maps for object localization. (2) The attention maps of visual transformers are semantic-agnostic (not distinguishable to object classes) and are not competent to semantic-aware localization, Fig.~\ref{fig:Flowchart_comparison}(b).

In this study, we propose the token semantic coupled attention map (TS-CAM), making the first attempt for weakly supervised object localization with visual transformer. 
TS-CAM introduces a semantic coupling structure with two network branches, Fig.~\ref{fig:Flowchart_comparison}(c), one performs semantic re-allocation using the patch tokens and the other generates semantic-agnostic attention map upon the class tokens.
Semantic re-allocation, with class-patch semantic activation, enables the patch tokens to be aware of object categories.
The semantic-agnostic attention map aims to capture long-distance feature dependency between patch tokens by taking the advantages of the cascaded self-attention modules in transformer.
TS-CAM finally couples the semantic-aware maps with the semantic-agnostic attention map for object localization, Fig. 

The contributions of this work are as follows:
\begin{itemize}
    \item We propose the token semantic coupled attention map (TS-CAM), as the first solid baseline for WSOL using visual transformer by leveraging the long-range feature dependency.
    
    \item We propose the semantic coupling module to combine the semantic-aware tokens with the semantic-agnostic attention map, providing a feasible way to leverage both semantics and positioning information extracted by visual transformer for object localization.
     
    \item TS-CAM achieves a substantial improvement over previous methods on two challenging WSOL benchmarks, fully exploiting the long-range feature dependency in the visual transformer.
   
\end{itemize}

\section{Related Work}
\label{sec:rel}
\noindent \textbf{Weakly Supervised Object Localization (WSOL)} aims to learn object localizations given solely image-level category labels. A representative study of WSOL is CAM~\cite{zhou2016learning}, which produces localization maps by aggregating deep feature maps using a class-specific fully connected layer. By removing the last fully connected layer, CAM can also be implemented by fully convolutional networks~\cite{Hwang2016selftransfer}.

Despite the simplicity and effectiveness of CAM-based methods, they suffer identifying small discriminative parts of objects. To improve the activation of CAMs, HaS~\cite{singh2017hide} and CutMix~\cite{singh2017hide} adopted adversarial erasing on input images to drive localization models focusing on extended object parts. ACoL~\cite{zhang2018adversarial} and ADL~\cite{choe2019attention} instead erased feature maps corresponding to discriminative regions and used adversarially trained classifiers to reconvert missed parts. SPG~\cite{zhang2018self} and I$^{2}$C~\cite{zhang2020inter} increased the quality of localization maps by introducing the constraint of pixel-level correlations into the network. DANet~\cite{xue2019danet} applied a divergent activation to learn complementary visual cues for WSOL. SEM~\cite{zhang2020rethinking} refined the localization maps by using the point-wise similarity within seed regions. GC-Net \cite{lu2020geometry} took geometric shapes into account and proposed a multi-task loss function for WSOL. 

Most of the above methods struggled to expand activation regions by introducing sophisticated spatial regularization techniques to CAM. However, they remain puzzled by the contradiction between image classification and object localization. As observed by the visualization approaches~\cite{Dissertion17,Understanding14}, CNNs tend to decompose an object into local semantic elements corresponding local receptive fields. Activating a couple of the semantic elements could bring good classification results. The problem about how to collect global cues from local receptive fields remains.

\noindent\textbf{Weakly Supervised Detection and Segmentation} are vision tasks closely related to WSOL.
Weakly supervised detection train networks to simultaneously perform image classification and instance localization~\cite{Wang_2018_CVPR,CMIL2019,WSOD2020}. Given thousands of region proposals, the learning procedure selects high-scored instances from bags while training detectors. In a similar way, weakly supervised segmentation trains classification networks to estimate pseudo masks which are further used for training the segmentation networks. To generate accurate pseudo masks,\cite{kolesnikov2016seed,Ahn_2018_CVPR,Huang_2018_CVPR,Wang_2018_CVPR} resorted to a region growing strategy. Meanwhile, some researchers investigated to directly enhance the feature-level activated regions~\cite{Lee_2019_CVPR,Wei_2018_CVPR}. Others accumulate CAMs by training with multiple phases~\cite{Jiang_2019_ICCV}, exploring boundary constraint~\cite{Chen_2020_ECCV}, leveraging equivalence for semantic segmentation~\cite{wang2020self}, and mining cross-image semantics~\cite{Sun_2020_ECCV} to refine pseudo masks.

Similar to WSOL, many weakly supervised detection and segmentation approaches are prone to localize object parts instead of full object extent. There is a requirement to explore new classification models to solve the partial activation problem in a systematic way.

\noindent\textbf{Long-Range Feature Dependency.} 
CNNs produce a hierarchical ensemble of local features with different reception fields. Unfortunately, most CNNs~\cite{VGG2014,ResNet2016} are good at extracting local features but experience difficulty to capture global cues.

To alleviate such a limitation, one solution is to utilize pixel similarity and global cues to refine  activation maps~\cite{Wang_2018_CVPR,wang2020self,zhang2020inter,zhang2020rethinking}.
Cao \etal~\cite{Cao_2019_ICCV} found that the global contexts modeled by non-local networks are almost the same for query positions and thereby proposed NLNet~\cite{Wang_2018_CVPR} with SENet~\cite{Hu_2018_CVPR} for global context modeling. MST~\cite{song_2019_nips_learnable} proposed the learnable tree filter to capture the structural property of minimal spanning tree to model long-range dependencies. The other solution is the attention mechanism~\cite{NonLocal2018,SASA2019}. The non-local operation~\cite{NonLocal2018} was introduced to CNNs in a self-attention manner so that the response at each position is a weighted sum of the features at all (global) positions. SASA~\cite{SASA2019} verified that self-attention is an effective stand-alone layer for CNNs. Relation Networks~\cite{RelationNet19} proposed to process a set of objects simultaneously through interaction between their features and geometry, allowing modeling the spatial relations between objects. Recent studies introduced a cascaded self-attention mechanism in the transformer model to capture long-range feature dependency~\cite{VT2020,DeiT2020,T2TViT2021}.


\section{Methodology}
\label{sec:method}
In this section, we first give the preliminaries for visual transformer. We then introduce the TS-CAM method. We finally analyze TS-CAM by feature visualization and feature dependency quantification.

\begin{figure*}[t]
    \centering
    \includegraphics[width=1.\linewidth]{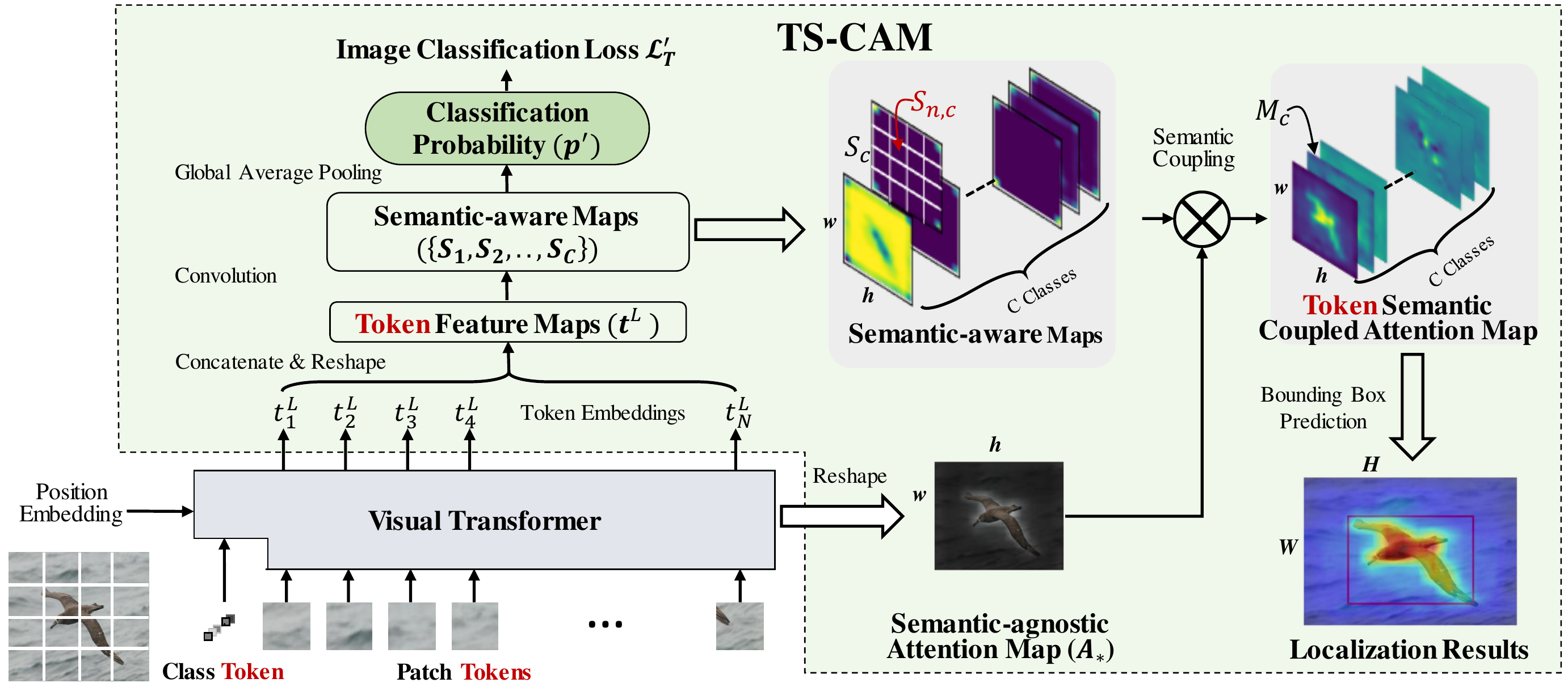}
    \caption{TS-CAM framework, which consists of a visual transformer for feature extraction, a semantic re-allocation branch and a semantic coupling module. Note that there is no gradient back-propagation along the semantic re-allocation branch.}
    \label{fig:Framework}
\end{figure*}

\subsection{Preliminaries}

For visual transformer~\cite{vit2020}, an input image $x$ of $W\times H$ resolution is divided to $w\times h$ patches, where $w=W/P, h=W/P$ and $P$ denotes the width/height of a patch. The divided patches are flattened and linearly projected to construct $N=w\times h$ patch tokens $\{t^0_n\in \mathbb{R}^{1\times D}, n=1,2,...,N\}$ and a class token $t_*^0 \in \mathbb{R}^{1\times D}$,  Fig.~\ref{fig:Framework}. D stands for the dimension of each token embedding. The class token $t_*^0$ is learnable with random initialization. Each token is added with a learnable position embedding in an element-wised manner. These tokens are fed to $L$ cascaded transformer blocks, each of which consists of a multi-head self-attention layer and a Multilayer Perceptron (MLP) block.

Denote $t_n^l$ and $t_*^l$ as the $n${\color{black}-}th patch token and the class token of the $l${\color{black}-}th transformer block, respectively. 
The last embedded class token $t_*^L$ is fed to an MLP block to predict the classification probability, as
\begin{equation}
    p = {\rm Softmax}\big({\rm MLP}(t_*^L)\big), 
    \label{eq:predict}
\end{equation}
where $p \in \mathbb{R} ^{1\times C}$ and $C$ denotes the number of classes. $p_c$ denotes the predicted probability to class $c$. MLP$(\cdot)$ denotes the classification function implemented by the MLP block.
Denote the ground-truth label for image $x$ as $y \in \{1,2,...,C\}$, the classification loss function is defined as
\begin{equation} \label{eq:loss_vit}
        {\cal L}_{T} = -\log p_{y},
\end{equation}
which is used to train the visual transformer.

\subsection{TS-CAM}

We propose the TS-CAM method to generate semantic-aware localization maps upon the trained visual transformer, Fig.~\ref{fig:Framework}. In the visual transformer, however, only the class token is semantic-aware while the patch tokens are semantic-agnostic. To fulfill semantic-aware localization, we introduce the semantic re-allocation branch to transfer semantics from the class token to patch tokens and generate semantic-aware maps. Such semantic-aware maps are coupled with the semantic-agnostic attention maps to generate the semantic-aware localization maps.

\textbf{Semantic Re-allocation.}
Visual transformer uses the class token to predict image categories (semantics), while using semantic-agnostic patch tokens to embed object spatial locations and reflects feature spatial dependency. To generate semantic-aware patch tokens, we propose to re-allocate the semantics from the class token $t_*^L$ to the patch tokens $\{t_1^L, t_2^L, ..., t_N^L\}$. 

As shown in Fig. \ref{fig:Framework}, patch token embeddings of the $L${\color{black}-}th visual transformer block are concatenated and transposed as $\mathbf{t}^L \in \mathbb{R}^{D\times N}$. They are then reshaped to token feature maps ${\mathbf{t}}^L\in \mathbb{R}^{D\times w\times h}$, where ${\mathbf{t}}^L_d, d\in \{1,2,...,D\}$ denotes the $d${\color{black}-}th feature map. The semantic aware map $S_{c}$ of class $c$ is calculated by convolution as
\begin{equation} \label{eq:semantic_activation_mapd}
     S_{c}= \sum_d {{\mathbf{t}}^L_{d}} * k_{c,d},
\end{equation}
where $k \in \mathbb{R}^{C\times D \times 3 \times 3}$ {\color{black}denotes} the convolution kernel and $k_{c,d}$ is a $3\times 3$ kernel map indexed by $c$ and $d$. $*$ is the convolution operator. To produce semantic-aware maps, the loss function defined in Eq.~\ref{eq:loss_vit} is updated to
\begin{equation} \label{eq:loss_ts_cam}
    \begin{split}
        {\cal L'}_{T} &= {-\log p'_{y}} \\
            &= -\log \frac
                {\exp\big({\sum_n S_{n,y}/N}\big)}
                {\sum_c \exp\big({\sum_n S_{n,c}/N}\big)},
    \end{split}
\end{equation}
where $S_{n,c}$ is the semantic of the $n${\color{black}-}th patch token for class $c$.
While optimizing Eq. \ref{eq:predict} allocates the semantics to class token $t_*^L$, minimizing Eq. \ref{eq:loss_ts_cam} re-allocates the semantics to patch tokens $\{t_1^L, t_2^L, ..., t_N^L\}$, generating semantic-aware maps for WSOL.

\textbf{Semantic-agnostic Attention Map.}
To fully exploit the long-range feature dependency of visual transformer, we propose to aggregate the attention vectors of the class token to generate the semantic-agnostic attention map.
Denote $\mathbf{t}^l \in \mathbb{R}^{(N+1)\times D}$ as the input of transformer block $l$, which is calculated by concatenating the embeddings of all tokens. In the self-attention operation in the $(l)${\color{black}-}th transformer block, the embedded tokens $\Tilde{\mathbf{t}}^{l}$ is computed as
\begin{equation} \label{eq:self_attention}
    \begin{split}
        \Tilde{\mathbf{t}}^{l} &= {\rm Softmax}\bigg((\mathbf{t}^l\theta^{l}_q)(\mathbf{t}^{l}\theta^{l}_k)^{\top}/\sqrt{D}\bigg) (\mathbf{t}^l\theta^{l}_v) \\
        &= A^l(\mathbf{t}^l\theta^{l}_v),
    \end{split}
\end{equation}
where $\theta^{l}_q$, $\theta^{l}_k$ and $\theta^{l}_v$ respectively denote parameters of the linear transformation layers of self-attention operation in $(l)$-th transformer block.
$\top$ is a transpose operator. $A^l$ is the attention matrix
while $A_*^l \in \mathbb{R}^{1 \times (N+1)}$ is the attention vector of the class token. In the multi-head attention layer where $K$ heads are considered, $D$ in Eq. \ref{eq:self_attention} is updated as $D'$, where $D'=D / K$. $A_*^l$ is then updated as the average of attention vectors from $K$ heads.

Eq.~\ref{eq:self_attention} implies that $A_*^l$ records the dependency of the class token to all tokens by the matrix multiplication operation. Eq.~\ref{eq:self_attention} implies that the embedding $\Tilde{t}_*^{l}$ of class token of the self-attention operation is calculated by multiplying its attention vector $A_*^l$ with the embedding $\mathbf{t}^{l}$ in the $(l)$-th transformer block. $\Tilde{t}_*^{l}$ is therefore able to ``see'' all patch tokens, where $A_*^l$ implies how much attention is paid on each token. When Eq. \ref{eq:loss_ts_cam} is optimized, the attention vector $A_*^l$ is driven to focus on object regions ($e.g.,$ long-range features of semantic correction) for image classification. The final attention vector $A_*$ is defined as
\begin{equation} \label{eq:attention_map}
    A_* = \frac{1}{L} \sum_l A_*^l,
\end{equation}
which aggregates attention vectors ($A_*^l$) and
collects feature dependency from cascaded transformer blocks to indicate full object extent.

\textbf{Semantic-Attention Coupling.}
As the attention vector $A_*$ is semantic-agnostic, we use an element-wise multiplication to couple it with the semantic-aware maps to obtain the semantic-coupled attention map $M_c$ for each class $c$, Fig.~\ref{fig:Framework}.
The coupling procedure is formulated as
\begin{equation} \label{eq:ts-cam}
    M_c = \Gamma^{w\times h} (A_*) \otimes S_c,
\end{equation}
where $\otimes$ denotes an element-wise multiplication and addition operations. $\Gamma^{w\times h}(\cdot)$ denotes the reshape function which converts the attention vector ($\mathbb{R}^{1\times N}$) to the attention map ($\mathbb{R}^{w\times h}$). $M_c$ is up-sampled to a semantic-aware localization map, which is used for object bounding box prediction with a thresholding approach~\cite{zhang2018self}. 

\begin{figure}[t]
    \centering
    \includegraphics[width=1.\linewidth]{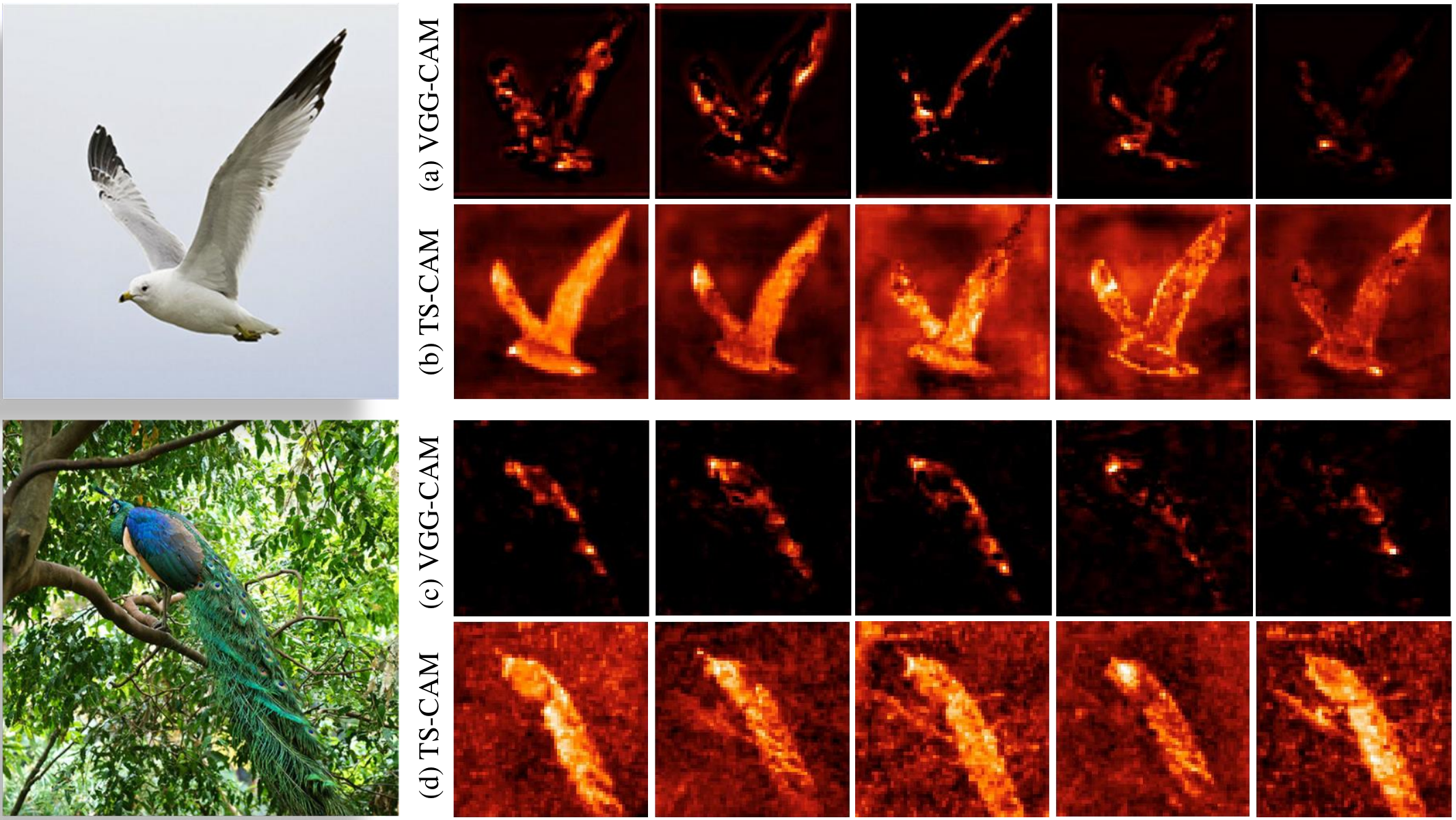}
    \caption{Comparison of feature maps of VGG-CAM and TS-CAM. In the feature maps of VGG-CAM, local discriminative regions are activated, while in TS-CAM, the activated features show long-range dependencies.}
    \label{fig:vis_compare_feature}
\end{figure}

\begin{figure}[t]
    \centering
    \includegraphics[width=1.0\linewidth]{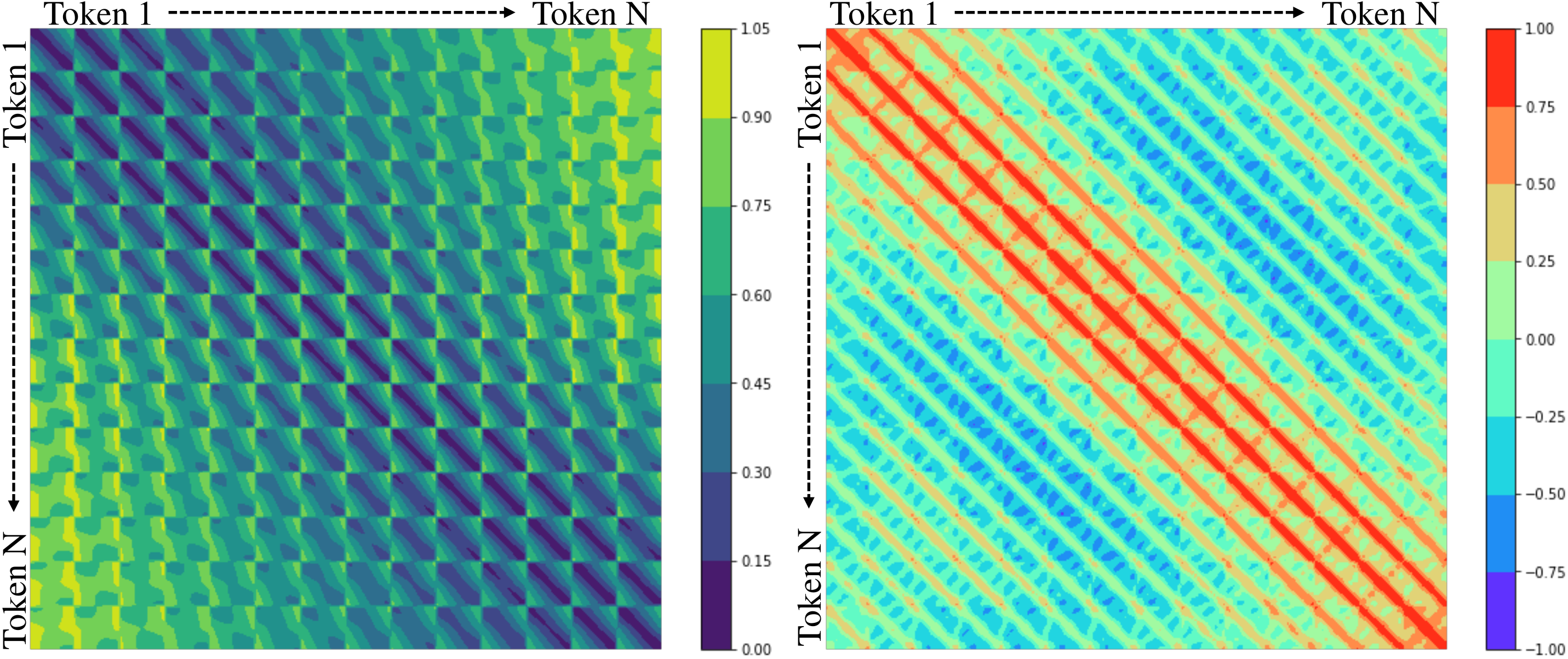}
    \caption{{\bf Left}: Distances among input patch tokens. Each row/column represents the Euclidean distances between a patch token with all patch tokens. Warmer/colder color means longer/short ranges. {\bf Right}: Similarities between position embeddings of patch tokens. Each row/column represents the cosine similarities between position  embedding of a token with all position embeddings. }
    \label{fig:vis-position-long-range}
\end{figure}

\subsection{Analysis of TS-CAM}
\textbf{Feature Visualization.} In Fig. \ref{fig:vis_compare_feature}, we visualize the feature maps of CNN and TS-CAM. The feature maps are randomly selected from different blocks/layers. It can be seen that the feature maps of CNN tend to activate local discriminative regions. In contrast, the feature maps of TS-CAM are able to activate full object extent.

\textbf{Long-Range Feature Dependency.}
In Fig.~\ref{fig:vis-position-long-range} right, we quantify the feature dependency ranges by using the cosine similarity between patch token embeddings (features). It can be seen that not only on the diagonal lines but also on the first and last rows/columns, the similarity values are significant despite the larger distance between them, Fig.~\ref{fig:vis-position-long-range} left. The similarity matrix shows that long-range feature dependency is captured by TS-CAM.

\begin{figure*}[t]
    \centering
    \includegraphics[width=1.\linewidth]{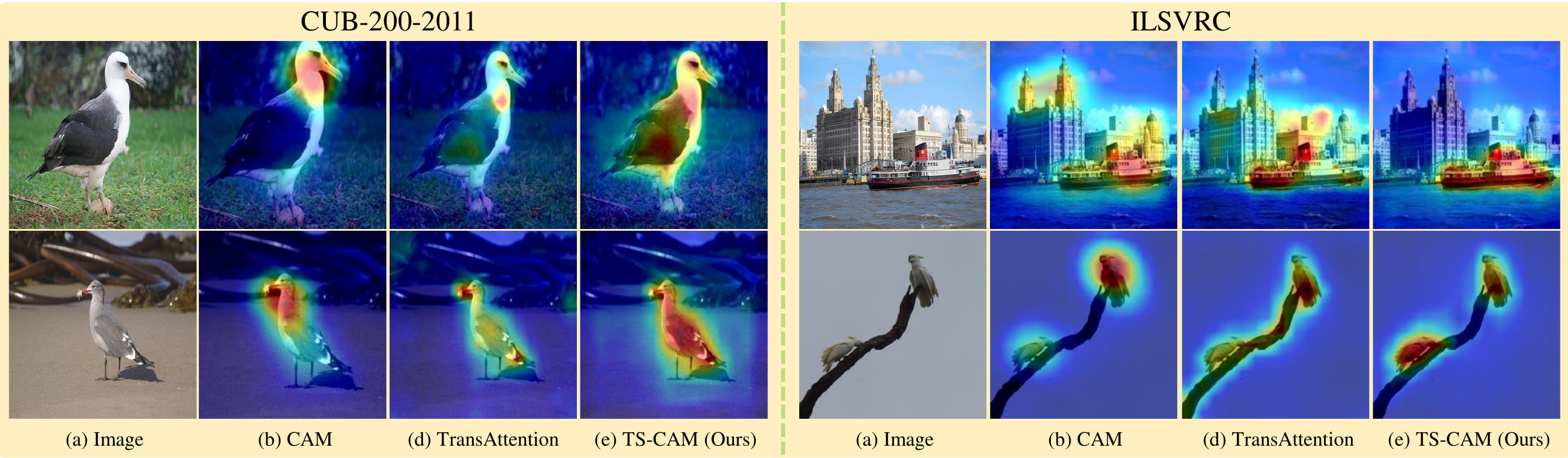}
    \caption{Visualization of localization maps on CUB-200-2011 and ILSVRC datasets. (a) Input Image. (b) Class Activation Map (CAM). (c) TransAttention:  Transformer-based Attention. (d) TS-CAM (ours). (Best Viewed in Color)}
    \label{fig:vis_cub_ilsvrc_loc}
\end{figure*}

\section{Experiments}
\label{sec:exps}

\subsection{Experimental Settings}
\noindent\textbf{Datasets.} 
TS-CAM is evaluated on two commonly used benchmarks, $i.e.$, CUB-200-2011~\cite{wah2011caltech} and ILSVRC~\cite{russakovsky2015imagenet}. 
CUB-200-2011 is a fine-grained bird dataset with $200$ different species, which is split into the training set of $5,994$ images and the test set of $5,794$ images.
In ILSVRC, there are around $1.2$ million images about $1,000$ categories for training and $50,000$ images for validation. The model is trained on the training set and evaluated on the validation set where the bounding box annotations are solely used for evaluation.

\noindent\textbf{Evaluation Metrics.}
Top-1/Top-5 classification accuracy (Top-1/Top-5 \emph{Cls. Acc}), Top-1/Top-5 localization accuracy (Top-1/Top-5 \emph{Loc. Acc}), and Localization accuracy with ground-truth class (Gt-Known \emph{Loc. Acc}) are adopted as evaluation metrics following baseline methods~\cite{russakovsky2015imagenet,zhang2018self,zhou2016learning}. 
For localization, a prediction is positive when it satisfies: the predicted classification is correct; the predicted bounding boxes have over 50$\%$ IoU with at least one of the ground-truth boxes. 
$Gt$-$Known$ indicates that it considers localization regardless of classification.

\noindent\textbf{Implementation Details.}
TS-CAM is implemented based on the Deit backbone~\cite{DeiT2020}, which is pre-trained on ILSVRC~\cite{russakovsky2015imagenet}. Each input image is re-scaled to 256$\times$256 pixels, and randomly cropped by 224$\times$224 pixels. We remove the MLP head, and add one convolution layer with kernel size 3$\times$3, stride 1, pad 1 with 200 output units (1000 units for ILSVRC). The newly added layer is initialized following He's approach~\cite{He2015DelvingDI}. When training WSOL models, we use AdamW~\cite{Loshchilov2019DecoupledWD} with $\epsilon$=1e-8, ${\beta}_1$=0.9 and ${\beta}_2$=0.99 and weight decay of 5e-4. On CUB-200-2011, the training procedure lasts 60 epochs with learning rate 5e-5 and batch-size 128. On ILSVRC dataset, training carries out 12 epochs with learning rate 5e-4 and batch-size 256. 

\begin{table}
\centering
\resizebox{.46\textwidth}{!}{
\begin{tabular}{l|c|c|c|c}
	\toprule
      \multirow{2}{*}{Methods} &\multirow{2}{*}{Backbone}  &\multicolumn{3}{|c}{Loc. Acc}   \\ 
      \cline{3-5}
       &  &Top-1 & Top-5&Gt-Known \\ \hline
       CAM~\cite{zhou2016learning} & GoogLeNet & 41.1 & 50.7 & 55.1  \\
       SPG~\cite{zhang2018self} & GoogLeNet & 46.7 & 57.2 & -  \\
       RCAM~\cite{zhang2020rethinking} & GoogleNet  & 53.0 & - & 70.0 \\
       DANet~\cite{xue2019danet} & InceptionV3 & 49.5 & 60.5 & 67.0  \\
       ADL~\cite{choe2019attention} & InceptionV3 & 53.0 & - & -  \\ \hline
       CAM~\cite{zhou2016learning} & VGG16 & 44.2 & 52.2 & 56.0  \\
       ADL~\cite{choe2019attention} & VGG16 & 52.4 & -& 75.4 \\
       ACoL~\cite{zhang2018adversarial} &VGG16 &45.9 & 56.5 & 59.3 \\
       DANet~\cite{xue2019danet} &VGG16 &52.5 & 62.0 & 67.7  \\
       SPG~\cite{zhang2018self} & VGG16 & 48.9 &57.2  & 58.9  \\
       I$^{2}$C~\cite{zhang2020inter} & VGG16 &56.0 &68.4 & -  \\
       MEIL~\cite{mai2020erasing} &VGG16 &57.5 & - & 73.8  \\
       RCAM~\cite{zhang2020rethinking}& VGG-16 & 59.0 & - & 76.3 \\ \hline
       TS-CAM (Ours) & Deit-S & {\bf 71.3} & {\bf 83.8} & {\bf 87.7} \\
	\bottomrule
\end{tabular}}
\vspace{.5em}
\caption{Comparison of TS-CAM with the state-of-the-art on the CUB-200-2011~\cite{wah2011caltech} test set.
}
\label{tab:cub_main}
\end{table}

\subsection{Performance}
\noindent \textbf{Main Results.}
Table~\ref{tab:cub_main} compares TS-CAM with other methods on the CUB-200-2011. 
TS-CAM with a Deit-S backbone~\cite{DeiT2020} outperforms the baseline methods on Top-1, Top-5, and Gt-Known metrics by a surprisingly large margin, yielding the localization accuracy of Top-1 $71.3\%$, and Top-5 $83.8\%$. Compared with the  state-of-the-art methods (RCAM \cite{zhang2020rethinking} and MEIL \cite{mai2020erasing}), it respectively achieves gains of 12.3$\%$ and 13.8$\%$ in terms of Top-1 \emph{Loc. Acc}. 
The left part of Fig.~\ref{fig:vis_cub_ilsvrc_loc} compares localization examples by CAM~\cite{zhou2016learning}, Transformer-based Attention, and TS-CAM on the CUB-200-2011. TS-CAM preserves global structures and covers more extent of objects. By solely utilizing the attention map from transformer structure, \emph{TransAttention} highlights most object parts, but fails to precisely localize full objects due to the lack of category semantics.

\begin{table}[!t]
\centering
\resizebox{.46\textwidth}{!}{
\begin{tabular}{l|c|c|c|c}
	\toprule
      \multirow{2}{*}{Methods} &\multirow{2}{*}{Backbone}  &\multicolumn{3}{|c}{Loc. Acc}  \\ 
      \cline{3-5}
       &  &Top-1 & Top-5&Gt-Known  \\ \hline
       Backprop~\cite{simonyan2013deep} & VGG16 & 38.9 & 48.5 & -  \\
       CAM~\cite{zhou2016learning} & VGG16 &42.8 &54.9 & 59.0  \\
       CutMix~\cite{yun2019cutmix} & VGG16 &43.5 &- & - \\
       ADL~\cite{choe2019attention} & VGG16 & 44.9 & -&- \\
       ACoL~\cite{zhang2018adversarial} &VGG16 & 45.8 &59.4 & 63.0  \\
       I$^{2}$C~\cite{zhang2020inter} & VGG16 &47.4 & 58.5 & 63.9  \\
       MEIL~\cite{mai2020erasing} &VGG16 &46.8 & - & -  \\
       RCAM ~\cite{zhang2020rethinking} & VGG-16 & 44.6 & - & 60.7 \\
       \hline
       CAM~\cite{zhou2016learning} &InceptionV3 &46.3 & 58.2 & 62.7 \\
       SPG~\cite{zhang2018self} & InceptionV3 & 48.6 & 60.0 & 64.7  \\
       ADL~\cite{choe2019attention} &InceptionV3 &48.7 & - & -  \\
       ACoL~\cite{zhang2018adversarial} & GoogLeNet & 46.7 & 57.4 & -  \\
       DANet~\cite{xue2019danet} &GoogLeNet & 47.5 & 58.3 &-  \\
       RCAM ~\cite{zhang2020rethinking} & GoogleNet & 50.6 & - & 64.4 \\
       MEIL~\cite{mai2020erasing} &InceptionV3 & 49.5 & - &-  \\
       I$^{2}$C~\cite{zhang2020inter} & InceptionV3 & 53.1 & 64.1 &{\bf 68.5}  \\
       GC-Net~\cite{lu2020geometry} & InceptionV3 & 49.1 & 58.1 & -  \\ \hline
       TS-CAM (Ours) & Deit-S & {\bf 53.4} & {\bf 64.3} & 67.6 \\
	\bottomrule
\end{tabular}
}
\vspace{0.5em}
\caption{Comparison of TS-CAM with state-of-the-art methods on the ILSVRC~\cite{russakovsky2015imagenet} validation set.
}
\label{tab:ilsvrc_main}
\end{table}

In Table~\ref{tab:ilsvrc_main}, we compare TS-CAM with its CNN counterparts (CAM) and the SOTAs on the localization accuracy by using tight bounding boxes on the ILSVRC.
TS-CAM respectively outperforms CAM on the VGG16~\cite{simonyan2014very} by $10.6\%$ and $9.4\%$ in terms of Top-1 \emph{Loc. Acc} and Top-5 \emph{Loc. Acc}. Compared with SOTAs with the VGG16 backbone~\cite{simonyan2014very}, TS-CAM outperforms by $\sim6\%$ and $\sim4\%$ in terms of Top-1 \emph{Loc. Acc} and Top-5 \emph{Loc. Acc}. Compared with $I^2C$, TS-CAM achieves performance gains of $6.0\%$ Top-1 \emph{Loc. Acc} and $5.8\%$ Top-5 \emph{Loc. Acc}, which are significant margins to the challenging problem.
%
%
Compared with SOTAs on the well-designed Inception V3~\cite{szegedy2016rethinking}, TS-CAM also achieves the best performance. Specifically, TS-CAM achieves performance gains of 7.1$\%$ and 6.1$\%$ in terms of Top-1 and Top-5 \emph{Loc. Acc} compared with CAM. 
Compared with I$^2$C which leverages pixel-level similarities across different objects to prompt the consistency of object features within the same categories, TS-CAM achieves comparable results with a cleaner and simpler pipeline.
The right half of Fig.~\ref{fig:vis_cub_ilsvrc_loc} illustrates examples of localization maps on ILSVRC.
CAM~\cite{zhou2016learning} tends to activate local discriminative regions and cannot retain the object structure well. Due to the lack of category semantics, \emph{TransAttention} activates almost the salient objects within images {\color{black}($e.g.$, the building in the first image and the branch in the second image)}.TS-CAM takes the advantage of self-attention mechanism in visual transformer and thus activates the full extent of objects. 

\begin{figure}[t]
    \centering
    \includegraphics[width=1.\linewidth]{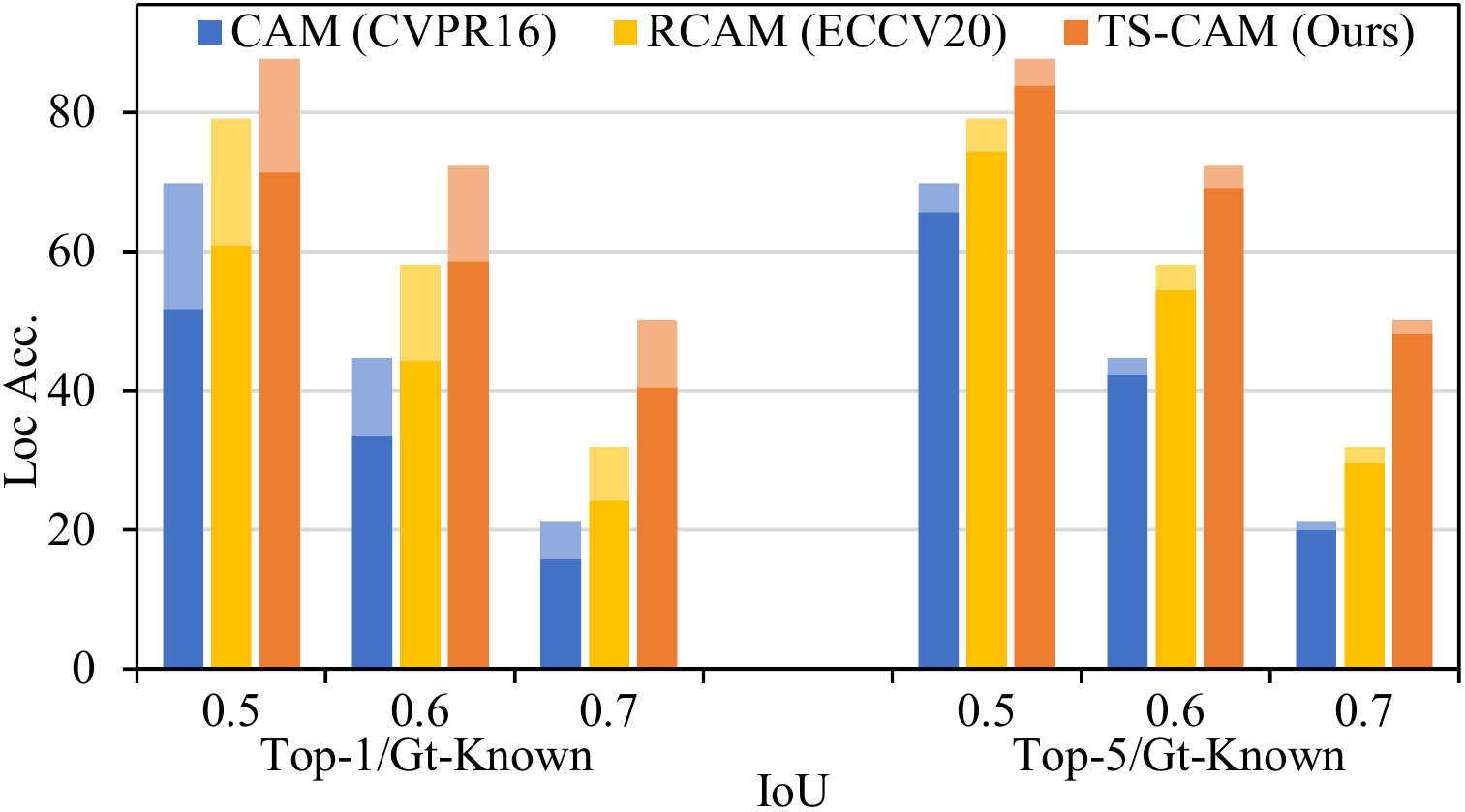}
    \caption{Comparison of localization accuracy under IoUs on CUB-200-2011~\cite{wah2011caltech}. Lighter color means Gt-Known \emph{Loc Acc}.}
    \label{fig:results_ious_cub}
\end{figure}

In Fig~\ref{fig:results_ious_cub}, we compare localization accuracy among variant CAM methods under different IoUs on CUB-200-2011~\cite{wah2011caltech}.
TS-CAM outperforms the CAM~\cite{zhou2016learning} and RCAM~\cite{zhang2020rethinking} under each IoU by large margins. 
In addition, TS-CAM achieves larger gains as IoU threshold increases, which indicates that the localization maps of our method cover the object extent accurately. 

\begin{table}
\centering
\resizebox{.46\textwidth}{!}{
\begin{tabular}{l|c|cc|c}
	\toprule
    \multirow{2}{*}{Methods} & Image  & $\# $Params & MACs  & Top-1 Loc Acc. \\
     & size & (M) & (G) & $\%$ \\
    \hline
    VGG16-CAM & $224^2$ & 19.6 & 16.3 & 44.2 \\
    GoogleNet-CAM & $224^2$ & 16.7  & 13.5  &  41.1 \\ \hline
    TS-CAM (Ours) & $224^2$  & 25.1 & 5.29 & 71.3    \\
	\bottomrule
\end{tabular}
}
\vspace{.5em}
\caption{Comparison of parameters and MACs. TS-CAM is implemented based on Deit-S~\cite{DeiT2020}. And Top-1 \emph{Loc Acc}. is evaluated on the CUB-200-2011 test set~\cite{wah2011caltech}. }
\label{tab:params_macs_analysis}
\end{table}

\noindent \textbf{Parameter Complexity.}
Under similar parameter complexity and computational cost overhead, TS-CAM (with 25.1M parameters and 5.29G MACs) respectively outperforms VGG16-CAM (with 19.6M parameters and 16.3G MACs) by 27.1$\%$ (71.3$\%$ vs. 44.2$\%$) and GoogleNet-CAM (with 16.7M parameters and 13.5G MACs)  by 27.2$\%$ (71.3$\%$ vs. 41.1$\%$) in Table~\ref{tab:params_macs_analysis}. 

\begin{table}
\centering
\resizebox{.46\textwidth}{!}{
\begin{tabular}{l|ccc|ccc}
	\toprule
    \multirow{2}{*}{Methods} &\multicolumn{3}{|c}{ILSVRC($\%$)}  &\multicolumn{3}{|c}{CUB-2011-200($\%$)} \\
     &M-Ins & Part & More &M-Ins & Part &More\\
    \hline
    VGG16 &10.65 &3.85 &9.58 &- &21.91 &10.53 \\
    InceptionV3 &10.36 &3.22 &9.49 &- &23.09&5.52 \\ \hline
     TS-CAM (Ours) &9.13  & 3.78 &7.65 &-  &6.30  &2.85  \\
	\bottomrule
\end{tabular}
}
\vspace{0.5em}
\caption{Localization error statistics.}
\label{tab:err_analysis_cub_ilsvrc}
\end{table}

\noindent \textbf{Error Analysis.}
To further reveal the effect of TS-CAM, we categorize the localization errors into five: classification error (Cls), multi-instance error (M-Ins), localization part error (Part), localization more error (More), and others (OT). \emph{Part} indicates that the predicted bounding box only cover the parts of object, and IoU is less than a certain threshold. \emph{More} indicates that the predicted bounding box is larger than the ground truth bounding box by a large margin. Each metric calculates the percentage of images belonging to corresponding error in the validation/test set. Table~\ref{tab:err_analysis_cub_ilsvrc} lists localization error statistics of \emph{M-Ins}, \emph{Part}, and \emph{More}. TS-CAM effectively reduces the \emph{M-Ins}, \emph{Part} and \emph{More} errors on both benchmarks, which indicates more accurate localization maps. 
For CUB-200-2011, TS-CAM significantly reduces both \emph{Part} and \emph{More}-type errors by $\sim17\%$ and $\sim3\%$ compared with CAM~\cite{zhou2016learning} on the basis of well-designed Inception V3\footnote{Please refer to supplementary materials for detailed analysis and definitions of each metric.}. 

\begin{figure*}[t]
    \centering
    \includegraphics[width=1.\linewidth]{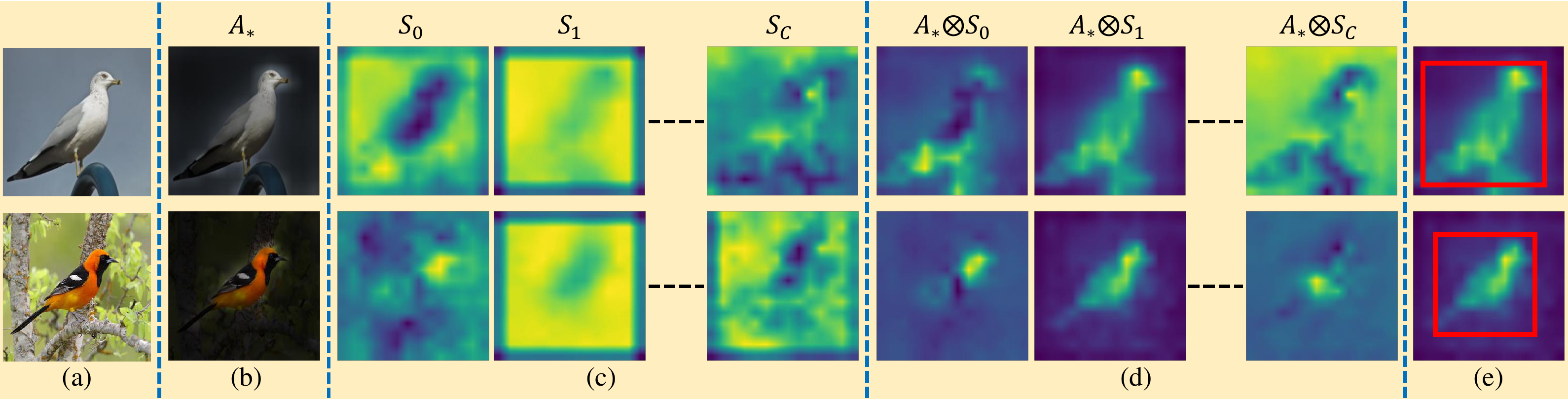}
    \caption{Visualization of semantic-attention coupling. (a) Input image. (b) Semantic-agnostic attention Map. (c) Token semantic-aware maps. (d) Token semantic coupled attention maps. (e) Localization results. (Best viewed in color)}
    \label{fig:vis_attn_semantic_couple}
\end{figure*}

\subsection{Ablation Study}
Using Deit-S as the backbone, we conduct ablation studies to verify the components in TS-CAM. In Table~\ref{tab:cub_baseline}, we evaluate the performance of the TS-CAM on CUB-200-2011 and observed significant improvements by TS-CAM components. \emph{TransAttention} solely uses semantic-agnostic attention map ($A_{*}$) for object localization, while \emph{TransCAM} solely uses semantic-aware maps ($S_{c}$). $A_{*}$ and $S_{c}$ are respectively generated by Eq. 6 and Eq. 3, and are illustrated in Fig. 3. Specifically, TS-CAM obtains gains of $12.4\%$, $14.1\%$, and $14.8\%$ in terms of Top-1, Top-5 and Gt-Known \emph{Loc. Acc} compared with \emph{TransAttention}. Using the semantic-aware map, \emph{TransCAM} struggles from distinguishing an object from the background due to the destruction of topology. Taking advantage of both modules, TS-CAM generates semantic-aware localization maps by coupling semantic-agnostic attention from transformer and token semantics from the classifier.

Table~\ref{tab:ilsvrc_baseline} shows the results on ILSVRC validation set with different configurations. Following CAM~\cite{zhou2016learning}, \emph{TransCAM} only utilizes the token semantic-aware map from the classifier to capture the object localization. Since the topology of input image is destroyed, \emph{TransCAM} cannot generate structure-preserving activation maps and thus cannot differentiate the objects from background. It obtains a significant performance degradation compared with TS-CAM and CAM. \emph{TransAttention} achieves 10.4$\%$ and 12.4$\%$ performance degradation compared TS-CAM in terms of Top-1 and Top-5 \emph{Loc. Acc}. The class-agnostic features puzzled \emph{TransAttention} toward false localization, Fig.~\ref{fig:Flowchart_comparison}(b).

\begin{figure}[t]
    \centering
    \includegraphics[width=1.0\linewidth]{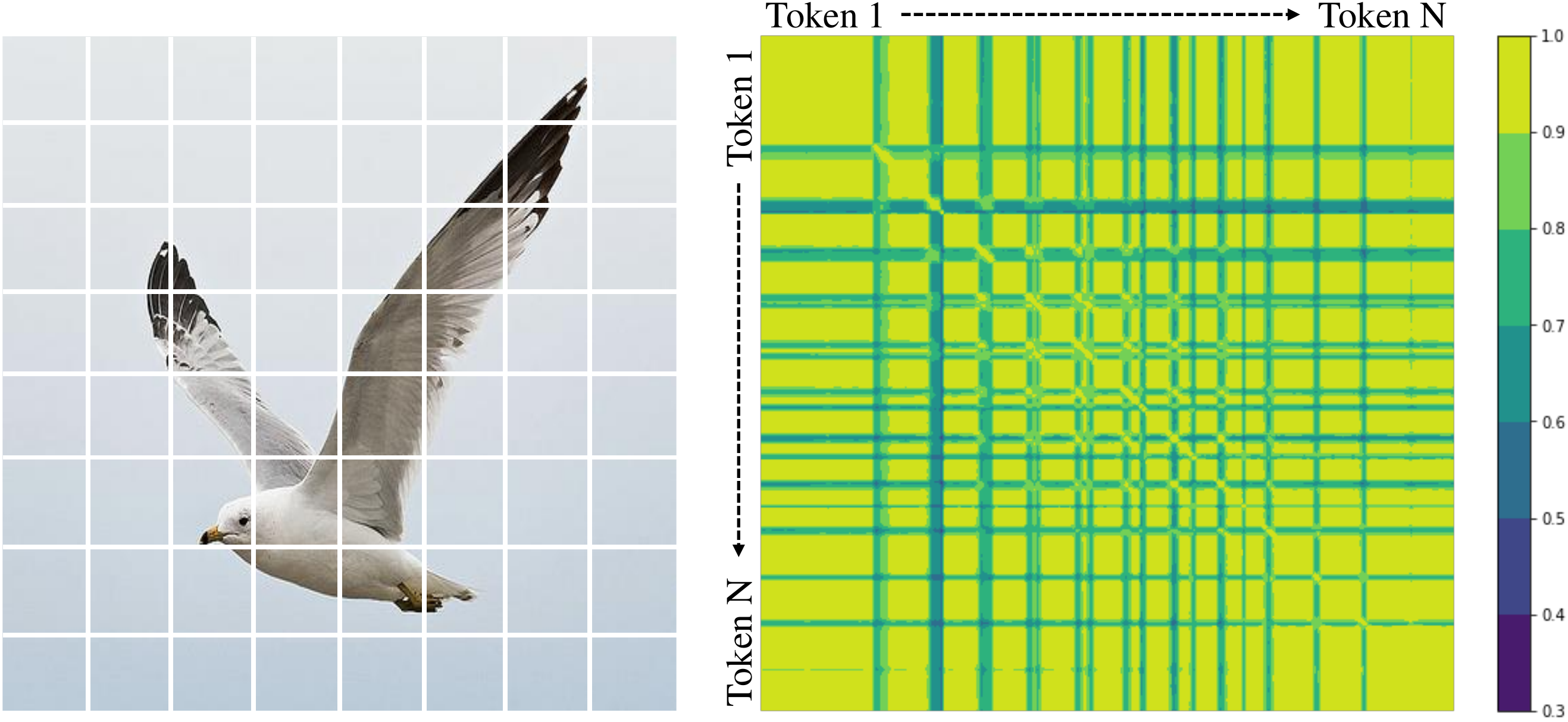}
    \caption{{\bf Left}:Input patch tokens. {\bf Right}: Visualization of the similarity matrix for patch token embeddings. Each row/column represents the cosine similarities between a patch token embedding with all patch token embeddings. }
    \label{fig:feature_similarity}
\end{figure}


\textbf{Why attention instead of activation?} The reasons are two folds: (1) Visual transformer leverages embeddings of a low-resolution class token for image classification and can not produce high-resolution CAM. (2) The semantic-aware maps in TS-CAM, which are calculated by re-allocating the semantics from the class token to patch tokens, fail to discriminatively activate the object regions. By visualizing the similarities among patch token embeddings $\{t^L_1,...,t^L_N\}$ in Fig.~\ref{fig:feature_similarity} right, we observed that the patch token embeddings are similar with each other, which implies that the activation results (semantic-aware activation maps) generated by these embeddings experience difficulty to discriminate objects from their backgrounds, Fig.~\ref{fig:vis_attn_semantic_couple}(c). 


\begin{table}
\centering
\resizebox{.46\textwidth}{!}{
\begin{tabular}{l|c|c|c|c}
	\toprule
      \multirow{2}{*}{Methods} &\multirow{2}{*}{Backbone}  &\multicolumn{3}{|c}{Loc Acc.}   \\ 
      \cline{3-5}
       &  &Top-1 & Top-5&Gt-Known \\ \hline
       \multirow{2}{*}{CAM~\cite{zhou2016learning}} & VGG-16 & 44.2 & 52.2 & 58.0  \\
       & GoogleNet & 41.2 & 51.7 & 55.1 \\ \hline
       TransAttention & Deit-S & 58.9 & 69.7 & 73.0  \\ \hline
       TransCam & Deit-S & 17.7 & 18.3 & 18.3  \\ \hline
       TS-CAM (Ours) & Deit-S & 71.3& 83.8 & 87.8 \\
	\bottomrule
\end{tabular}}
\vspace{.5em}
\caption{Ablation studies of TS-CAM components on the CUB-200-2011 test set~\cite{wah2011caltech}.
}
\label{tab:cub_baseline}
\end{table}

\begin{table}
\centering
\resizebox{.46\textwidth}{!}{
\begin{tabular}{l|c|c|c|c}
	\toprule
      \multirow{2}{*}{Methods} &\multirow{2}{*}{Backbone}  &\multicolumn{3}{|c}{Loc Acc.}   \\ 
      \cline{3-5}
       &  &Top-1 & Top-5&Gt-Known \\ \hline
       \multirow{2}{*}{CAM} & VGG-16 & 42.8 & 54.9 & 59.0  \\
       & InceptionV3 & 46.3 & 58.2 & 62.7 \\ \hline
       TransAttention & Deit-S & 43.0 & 51.9 & 54.7  \\ \hline
       
       TransCam & Deit-S & 34.9 & 42.9 & 46.0  \\ \hline
       TS-CAM (Ours) & Deit-S & 53.4 & 64.3 & 67.6 \\
	\bottomrule
\end{tabular}}
\vspace{.5em}
\caption{Ablation study of TS-CAM components on the ILSVRC validation set~\cite{russakovsky2015imagenet}.
}
\label{tab:ilsvrc_baseline}
\end{table}


\section{Conclusion}
\label{sec:con}

 We proposed the token semantic coupled attention map (TS-CAM) for weakly supervised object localization.
 TS-CAM takes full advantage of the cascaded self-attention mechanism in the visual transformer for long-range feature dependency extraction and object extent localization.
 To solve the semantic agnostic issue of the patch tokens, we proposed to re-allocate category-related semantics for patch tokens, enabling each of them to be aware of object categories.
 We proposed the semantic coupling strategy to fuse the patch tokens with the semantic-agnostic attention map to achieve semantic-aware localization results.
 Experiments on the ILSVRC/CUB-200-2011 datasets show that TS-CAM significantly improved the WSOL performance, in striking contrast with its CNN counterpart (CAM).
  As the first and solid baseline with transformer, TS-CAM provides a fresh insight to the challenging WSOL problem.

{\small
\bibliographystyle{ieee_fullname}
\bibliography{egbib}
}


\appendix
\newpage

\section{Appendix}
\subsection{Additional Error Analysis}
\label{sec:sup}
\begin{table*}
\centering
\setlength{\tabcolsep}{1.6mm}{
\begin{tabular}{l|ccccc|ccccc}
	\toprule
    \multirow{2}{*}{Methods} &\multicolumn{5}{|c}{CUB-2011-200($\%$)} &\multicolumn{5}{|c}{ILSVRC($\%$)}  \\
     &Cls&M-Ins & Part & More & OT & Cls &M-Ins & Part &More & OT\\
    \hline
    VGG16 & 23.28 &- &21.91 &10.53 & 1.76 & 29.47 &10.65 &3.85 &9.58 & 0.20 \\
    InceptionV3 & 26.39&- &23.09&5.52 &0.64 & 26.73&10.36 &3.22 &9.49 & 0.20 \\ \hline
     TS-CAM (Ours) & 19.66 &-  &6.30  &2.85 & 0.01 & 25.39 & 9.13  & 3.78 &7.65 & 0.41 \\
	\bottomrule
\end{tabular}
\vspace{0.5em}
\caption{Localization error statistics.}
\label{tab:err_analysis_cub_ilsvrc_sup}
}
\end{table*}

\paragraph{Details in Error Analysis}
In this sub-section, the detailed definitions in Section 4.2 (Error Analysis) of the paper are given. Specifically, classification error (Cls), multi-instance error (M-Ins), localization part 
error (Part), localization more error (More), and others (OT) are respectively defined as follows.
\begin{itemize}
    \item \emph{Cls} indicates the predictions which are falsely classified.
    \item \emph{M-Ins} indicates that the prediction intersects with at least two ground-truth boxes, and IoG $>$ 0.3.
    \item \emph{Part} indicates that the predicted bounding box covers partial object, and IoP $>$ 0.5.
    \item \emph{More} indicates that the predicted bounding box is larger than the ground truth bounding box by a large margin, and IoG $>$ 0.7.
    \item \emph{OT} indicates other predictions that do not belong to the above mentioned cases.
\end{itemize}
where IoG and IoP are defined as intersection over ground truth box and intersection over predict bounding box, respectively (similar to \emph{IoU}(intersection over Union). The five cases defined above are mutually exclusive, Algorithm~\ref{alg:err_alg}. Each metric calculates the percentage of images belonging to corresponding error in the validation/testing set.

\begin{figure*}[t]
    \centering
    \includegraphics[width=1.\linewidth]{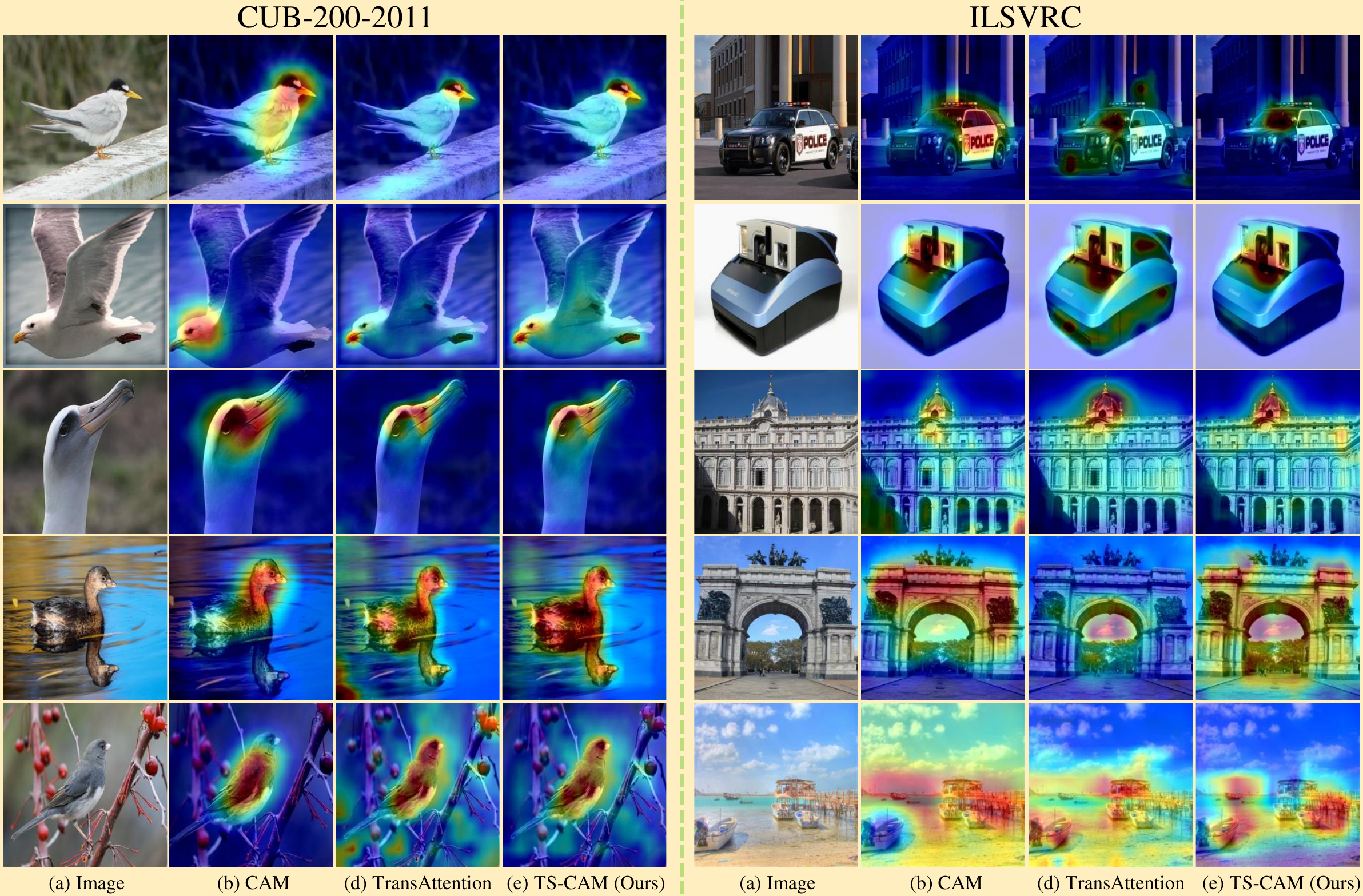}
    \caption{Failure cases on CUB-200-2011 and ILSVRC datasets. (a) Input Image. (b) Class Activation Map (CAM). (c) TransAttention:  Transformer-based Attention. (d) TS-CAM (ours). (Best Viewed in Color)}
    \label{fig:vis_cub_ilsvrc_err_loc_sup}
\end{figure*}

\begin{algorithm}[t]
 \label{alg:err_alg}  
\renewcommand{\algorithmicrequire}{ \textbf{Input:}}
\renewcommand\algorithmicensure { \textbf{Output:} }
\caption{ Error Analysis Algorithm.}
\begin{algorithmic}[1]
\Require  { Image Set ${\cal X}=\{x_i, i=1,2,...I\}$. For each image $x_i$, predicted bounding box $b_p \in \mathbb{R}^{1 \times 4 }$; ground truth bounding boxes $b_g\in \mathbb{R}^{Q \times 4 }$; predicted label ${\Tilde{y}} \in \{1,2,...,C\}$; ground truth label $y \in \{1,2,...,C\}$; maximum IoU between predictions and ground truth boxes $IoU_m$}
\Ensure  {Cls, M-Ins, Part, More, OT}
    \State {Set Cls = M-Ins = Part = More = OT = 0}
    \For {$x_i$ in $\cal X$}
        \State {Calculate the $IoG$ and $IoP \in \mathbb{R}^{Q \times 1} $}
        \State {$IoG_m = \max_q IoG$}
        \State {$IoP_m = \max_q IoP$}
        \If {$ {\Tilde{y}}(x_i) \neq y(x_i)$}
            \State {Cls = Cls + 1}
        \ElsIf {$IoU_m < 0.5$ }
            \If {$Count(IoG > 0.3) > 1 $}
                \State {M-Ins = M-Ins + 1 }
            \ElsIf {$IoP_m > 0.5$}
                \State {Part = Part + 1}
            \ElsIf {$IoG_m > 0.7$}
                \State {More = More + 1}
            \Else
                \State {OT = OT + 1} 
            \EndIf
        \EndIf
    \EndFor
    \State {Divide Cls, M-Ins, Part, More, OT by $I$.}
\end{algorithmic}
\end{algorithm}

Table.~\ref{tab:err_analysis_cub_ilsvrc_sup} shows the complete results of the error analysis. It can be seen that TS-CAM largely reduces the classification error on both CUB-2011-200 and ILSVRC datasets. Compared with CAM, TS-CAM reduces the error rate of $OT$ on CUB-2011-200 dataset and has comparable error rate of $OT$ on ILSVRC dataset.

\paragraph{Visualization of Failure Cases}

Fig.~\ref{fig:vis_cub_ilsvrc_err_loc_sup} shows the failure cases on CUB-200-2011 test set (left) and ILSVRC validation set (right). In these cases, TS-CAM focuses on parts of the textureless objects, or highlights semantic-related background regions, which are also observed in the results of CAM~\cite{zhou2016learning} and TransAttention.

\subsection{Performance}

\paragraph{Complete Performance Comparison}
\begin{table*}
\centering
\setlength{\tabcolsep}{2.8mm}{
\begin{tabular}{l|c|c|c|c|c|c}
	\toprule
      \multirow{2}{*}{Methods} &\multirow{2}{*}{Backbone}  &\multicolumn{3}{|c}{Loc. Acc} & \multicolumn{2}{|c}{Cls Acc.}  \\ 
      \cline{3-7}
       &  &Top-1 & Top-5&Gt-Known & Top-1 & Top-5 \\ \hline
       CAM~\cite{zhou2016learning} & GoogLeNet & 41.1 & 50.7 & 55.1 & 73.8 & 91.5 \\
       SPG~\cite{zhang2018self} & GoogLeNet & 46.7 & 57.2 & - & -& - \\
       RCAM~\cite{zhang2020rethinking} & GoogleNet  & 53.0 & - & 70.0 & 73.7 & - \\
       DANet~\cite{xue2019danet} & InceptionV3 & 49.5 & 60.5 & 67.0 & 71.2 & 90.6 \\
       ADL~\cite{choe2019attention} & InceptionV3 & 53.0 & - & -& 74.6 & -  \\ \hline
       CAM~\cite{zhou2016learning} & VGG16 & 44.2 & 52.2 & 56.0 & 76.6 & 92.5 \\
       ADL~\cite{choe2019attention} & VGG16 & 52.4 & -& 75.4 & 65.3 & - \\
       ACoL~\cite{zhang2018adversarial} &VGG16 &45.9 & 56.5 & 59.3 & 71.9 & - \\
       DANet~\cite{xue2019danet} &VGG16 &52.5 & 62.0 & 67.7 & 75.4 & 92.3 \\
       SPG~\cite{zhang2018self} & VGG16 & 48.9 &57.2  & 58.9 & 75.5 & 92.1  \\
       I$^{2}$C~\cite{zhang2020inter} & VGG16 &56.0 &68.4 & - & - & -  \\
       MEIL~\cite{mai2020erasing} &VGG16 &57.5 & - & 73.8  & 74.8 & - \\
       RCAM~\cite{zhang2020rethinking}& VGG-16 & 59.0 & - & 76.3 & 75.0 & - \\ \hline
       TS-CAM (Ours) & Deit-S & {\bf 71.3} & {\bf 83.8} & {\bf 87.7} & {\bf 80.3} & {\bf 94.8} \\
	\bottomrule
\end{tabular}
\vspace{.5em}
\caption{Comparison of TS-CAM with the state-of-the-art on the CUB-200-2011~\cite{wah2011caltech} test set.
}
\label{tab:cub_main_sup}
}
\end{table*}

Table~\ref{tab:cub_main_sup} shows more performance comparisons with state-of-the-art methods on CUB-2011-200 test set. In addition to the large improvement of localization accuracy, TS-CAM significantly outperforms the state-of-the-art methods for Top-1 and Top-5 $Cls.$ $Acc.$ by about 4\% and 2\% respectively. These results clearly demonstrate that TS-CAM takes the advantage of self-attention mechanism in visual transformer, which not only activates the full extent of objects but also obtains more discriminative image classifiers.


\begin{table*}[!t]
\centering
\setlength{\tabcolsep}{3.2mm}{
\begin{tabular}{l|c|c|c|c|c|c}
	\toprule
      \multirow{2}{*}{Methods} &\multirow{2}{*}{Backbone}  &\multicolumn{3}{|c}{Loc. Acc} & \multicolumn{2}{|c}{Cls Acc.}  \\ 
      \cline{3-7}
       &  &Top-1 & Top-5&Gt-Known & Top-1 & Top-5 \\ \hline
       Backprop~\cite{simonyan2013deep} & VGG16 & 38.9 & 48.5 & -  & - & - \\
       CAM~\cite{zhou2016learning} & VGG16 &42.8 &54.9 & 59.0 & 68.8 & 88.6 \\
       CutMix~\cite{yun2019cutmix} & VGG16 &43.5 &- & - & -& - \\
       ADL~\cite{choe2019attention} & VGG16 & 44.9 & -&-  & - & - \\
       ACoL~\cite{zhang2018adversarial} &VGG16 & 45.8 &59.4 & 63.0 & 67.5 & 88.0  \\
       I$^{2}$C~\cite{zhang2020inter} & VGG16 &47.4 & 58.5 & 63.9 & 69.4 & 89.3\\
       MEIL~\cite{mai2020erasing} &VGG16 &46.8 & - & - & 70.3 & - \\
       RCAM ~\cite{zhang2020rethinking} & VGG-16 & 44.6 & - & 60.7 & 68.7 & -\\
       \hline
       CAM~\cite{zhou2016learning} &InceptionV3 &46.3 & 58.2 & 62.7 & 73.3 & 91.8 \\
       SPG~\cite{zhang2018self} & InceptionV3 & 48.6 & 60.0 & 64.7 & 69.7 & 90.1\\
       ADL~\cite{choe2019attention} &InceptionV3 &48.7 & - & - & 72.8 & - \\
       ACoL~\cite{zhang2018adversarial} & GoogLeNet & 46.7 & 57.4 & - & 71.0 & 88.2 \\
       DANet~\cite{xue2019danet} &GoogLeNet & 47.5 & 58.3 & - & 63.5 & 91.4 \\
       RCAM ~\cite{zhang2020rethinking} & GoogleNet & 50.6 & - & 64.4 & 74.3 & - \\
       MEIL~\cite{mai2020erasing} &InceptionV3 & 49.5 & - &- & 73.3 & - \\
       I$^{2}$C~\cite{zhang2020inter} & InceptionV3 & 53.1 & 64.1 &{\bf 68.5} & 73.3 & 91.6  \\
       GC-Net~\cite{lu2020geometry} & InceptionV3 & 49.1 & 58.1 & - & \textbf{77.4} & \textbf{93.6} \\ \hline
       TS-CAM (Ours) & Deit-S & {\bf 53.4} & {\bf 64.3} & 67.6 & 74.3 & 92.1 \\
	\bottomrule
\end{tabular}
\vspace{0.5em}
\caption{Comparison of TS-CAM with state-of-the-art methods on the ILSVRC~\cite{russakovsky2015imagenet} validation set.
}\vspace{-1.5em}

\label{tab:ilsvrc_main_sup}
}
\end{table*}

Table~\ref{tab:ilsvrc_main_sup} shows the complete performance comparison with state-of-the-art methods on the ILSVRC validation set. It can be seen that TS-CAM outperforms the baseline method CAM with InceptionV3 backbone for Top-1 and Top-5 $Cls.$ $ Acc.$ by about 1.0\% and 0.3\% respectively. TS-CAM also outperforms the state-of-the-art methods based on VGG16~\cite{VGG2014} backbone for image classification. Compared with GC-Net \cite{lu2020geometry}, TS-CAM achieves worse classification performance but significantly better localization performance, which implies GC-Net suffering from poor localization ability while TS-CAM utilizing the long-range feature dependency of visual transformer and able to localize full object extent.


\begin{figure*}[h]
    \centering
    \includegraphics[width=1.\linewidth]{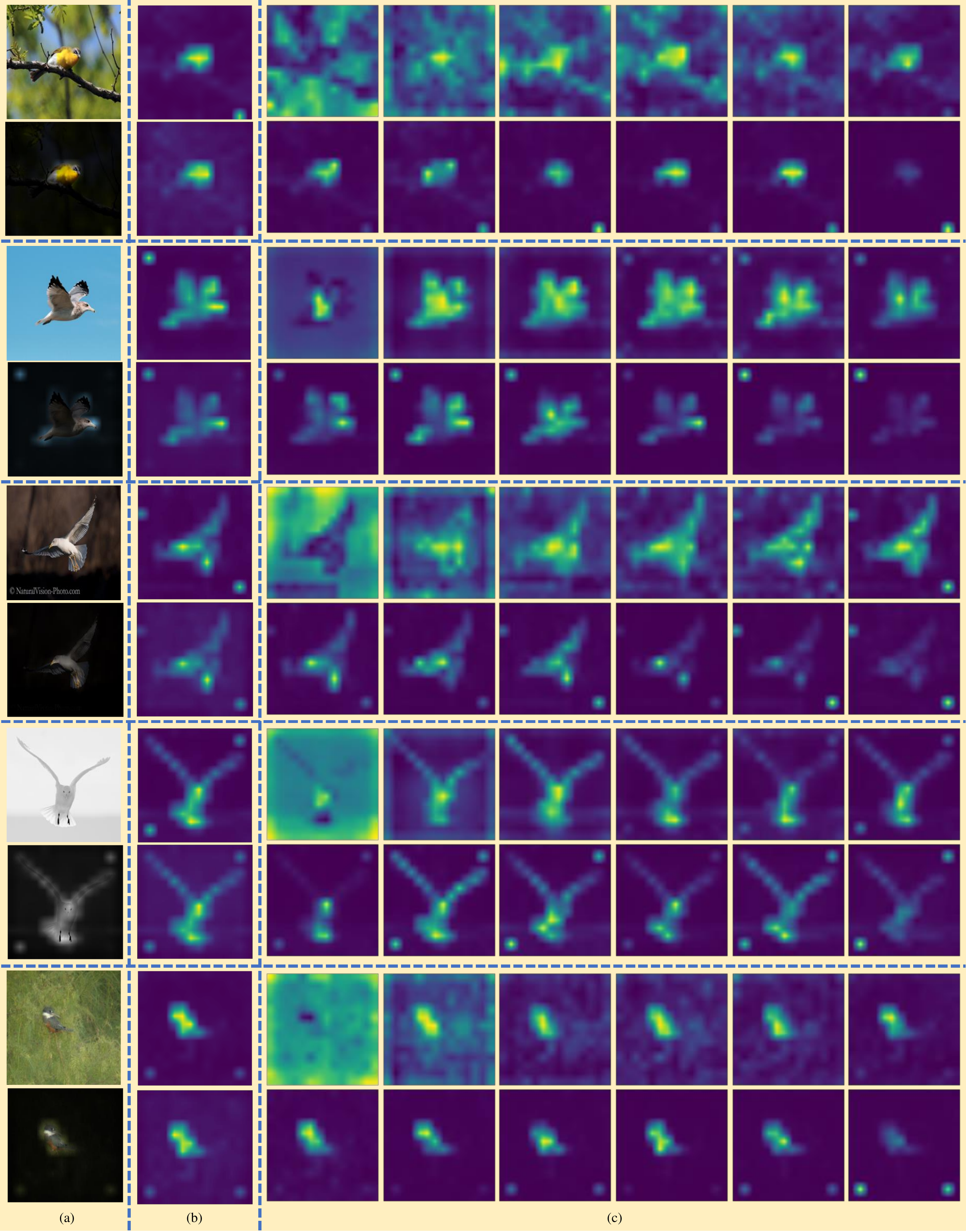}
    \caption{Visualization of attention maps on CUB-200-2011 test set. (a) Top: input image; Bottom: input image with $M_c$. (b) Top: semantic-agnostic attention map($A_*$); Bottom: semantic-aware localization map($M_c$). (c) Attention maps($A_*^l$) from different layers. (Best Viewed in Color)}
    \label{fig:vis_cub_attention_sup}
    \vspace{-0.5em}
\end{figure*}

\begin{figure*}[h]
    \centering
    \includegraphics[width=1.\linewidth]{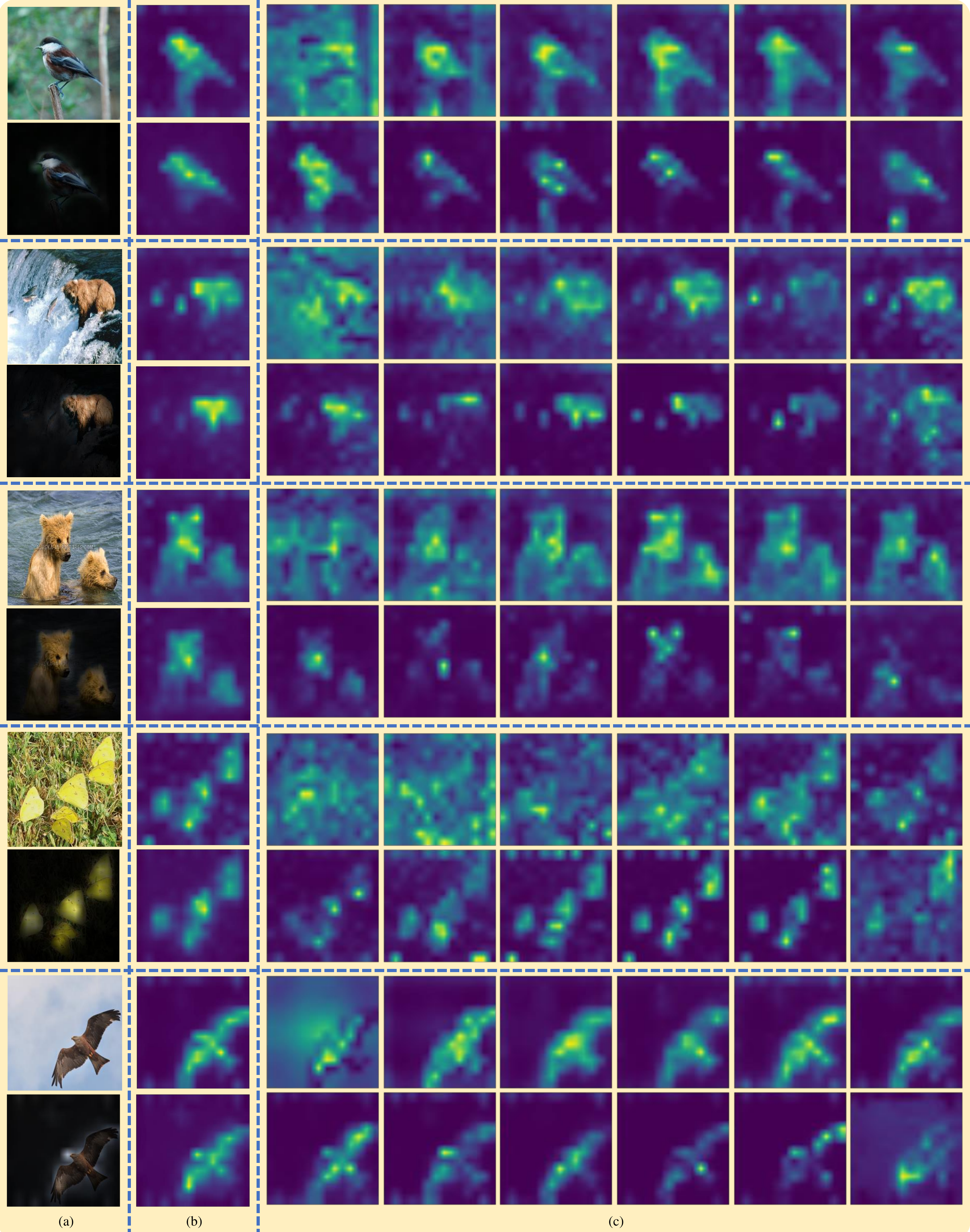}
    \caption{Visualization of attention maps on ILSVRC validation set. (a) Top: input image; Bottom: input image with $M_c$. (b) Top: semantic-agnostic attention map($A_*$); Bottom: semantic-aware localization map($M_c$). (c) Attention maps($A_*^l$) from different layers. (Best Viewed in Color)}
    \label{fig:vis_ilsvrc_attention_sup}
    \vspace{-0.5em}
\end{figure*}

\begin{figure*}[h]
    \centering
    \includegraphics[width=1.\linewidth]{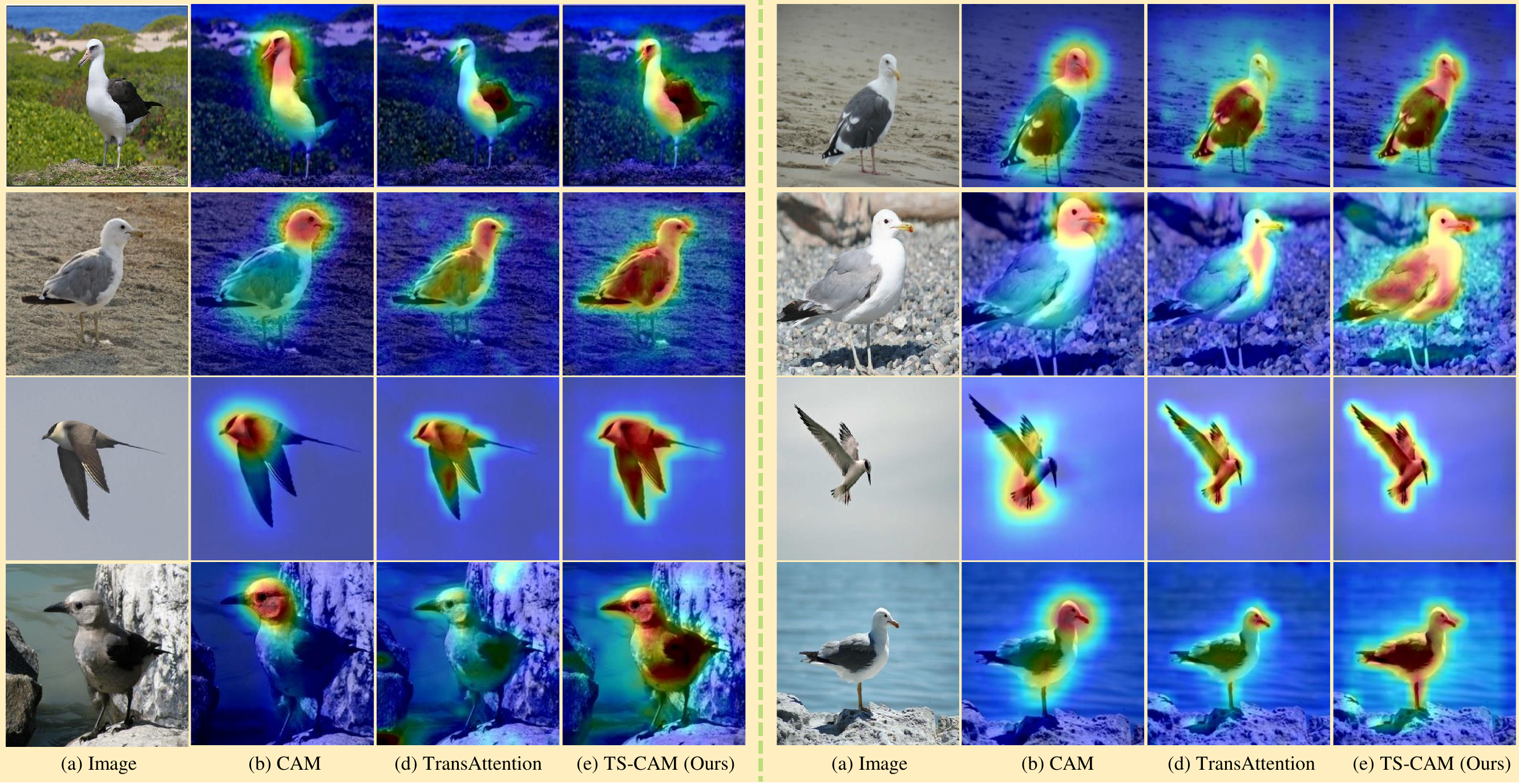}
    \caption{Visualization of localization maps on CUB-200-2011 test set. (a) Input Image. (b) Class Activation Map (CAM). (c) TransAttention:  Transformer-based Attention. (d) TS-CAM (ours). (Best Viewed in Color)}
    \label{fig:vis_cub_loc_sup}
    \vspace{-0.5em}
\end{figure*}

\paragraph{Ablation Study}
\begin{table*}
\centering
\setlength{\tabcolsep}{2.8mm}{
\begin{tabular}{l|c|c|c|c|c|c}
	\toprule
      \multirow{2}{*}{Methods} &\multirow{2}{*}{Backbone}  &\multicolumn{3}{|c}{Loc. Acc} & \multicolumn{2}{|c}{Cls Acc.}  \\ 
      \cline{3-7}
       &  &Top-1 & Top-5&Gt-Known & Top-1 & Top-5 \\ \hline
       FC &Deit-S &69.4 & 82.2 & 86.0  & 79.0 & 94.3 \\
       Conv1D &Deit-S &70.8 & 82.7 & 86.4  & 80.4 & 94.5 \\
       Conv2D(Ours) & Deit-S & {\bf 71.3} & {\bf 83.8} & {\bf 87.7} & {\bf 80.3} & {\bf 94.8} \\
	\bottomrule
\end{tabular}
\vspace{.5em}
\caption{Ablation results of TS-CAM with different classification layers on CUB-200-2011~\cite{wah2011caltech} test set. \emph{FC} means fully-connected layer, \emph{Conv1D} means 1D convolutional layer with kernel size 3 and \emph{Conv2D} means 2D convolutional layer with kernel size 3$\times$3.
}
\label{tab:cub_main_head_sup}
}
\end{table*}

\begin{table*}
\centering
\setlength{\tabcolsep}{2.8mm}{
\begin{tabular}{l|c|c|c|c|c|c}
	\toprule
      \multirow{2}{*}{Methods} &\multirow{2}{*}{Backbone}  &\multicolumn{3}{|c}{Loc. Acc} & \multicolumn{2}{|c}{Cls Acc.}  \\ 
      \cline{3-7}
       &  &Top-1 & Top-5&Gt-Known & Top-1 & Top-5 \\ \hline
       FC &Deit-S &31.5 &38.2 & 40.5  & 73.6 & 91.5 \\
       Conv1D &Deit-S & 49.8 & 60.1 & 63.3  & 73.7 & 91.5 \\
       Conv2D(Ours) & Deit-S & {\bf 53.4} & {\bf 64.3} & {\bf 67.6} & {\bf 74.3} & {\bf 92.1} \\
	\bottomrule
\end{tabular}
\vspace{.5em}
\caption{Ablation results of TS-CAM with different classification layers on ILSVRC~\cite{russakovsky2015imagenet} validation set. \emph{FC} means fully-connected layer, \emph{Conv1D} means 1D convolutional layer with kernel size 3 and \emph{Conv2D} means 2D convolutional layer with kernel size 3$\times$3.
}
\label{tab:ilsvrc_main_head_sup}
}
\end{table*}

\begin{table*}
\centering
\setlength{\tabcolsep}{2.8mm}{
\begin{tabular}{c|c|c|c}
	\toprule
      \multirow{2}{*}{$L$}  &\multicolumn{3}{|c}{Loc Acc.}  \\ 
      \cline{2-4}
          &Top-1 & Top-5&Gt-Known  \\ \hline
       8 & 65.2 & 75.8 & 79.0 \\
       9 & 68.5 & 80.0 & 83.6  \\
       10 & 70.2 & 81.2 & 85.5  \\
       11 & 71.2 & 83.7 & 87.7  \\
       12 & \textbf{71.3} & \textbf{83.8} & \textbf{87.7} \\
	\bottomrule
  \end{tabular}
\vspace{.5em}
\caption{Ablation results of TS-CAM when attention maps($A_*^l$) from different layers on CUB-200-2011~\cite{wah2011caltech} test set.
}
\label{tab:cub_ablation_attention_layer}
}
\end{table*}

\begin{table*}
\centering
\setlength{\tabcolsep}{2.8mm}{
\begin{tabular}{c|c|c|c}
	\toprule
      \multirow{2}{*}{$L$}  &\multicolumn{3}{|c}{Loc Acc.}  \\ 
      \cline{2-4}
          &Top-1 & Top-5&Gt-Known  \\ \hline
       8 & 49.4 & 59.9 & 63.2  \\
       9 & 50.8 & 61.6 & 65.0  \\
       10 & 52.2 & 63.3 & 66.7  \\
       11 & 52.8 & 64.0 & 67.5  \\
       12 & \textbf{53.4} & \textbf{64.3} & \textbf{67.6} \\
	\bottomrule
  \end{tabular}
\vspace{.5em}
\caption{Ablation results of TS-CAM when summing attention maps($A_*^l$) from different layers on ILSVRC~\cite{russakovsky2015imagenet} validation set. 
}
\label{tab:ilsvrc_ablation_attention_layer}
}
\end{table*}

In Table~\ref{tab:cub_main_head_sup} and Table~\ref{tab:ilsvrc_main_head_sup} , we make ablation study of the classification layer in Semantic Re-allocation, Section 3.2 in the paper. We replace the convolutional layer (\emph{Conv2D}) for semantic-aware map generation with a fully connected layer (\emph{FC}) or a 1D convolutional layer (\emph{Conv1D}). It can be seen that \emph{Conv2D} achieves better performance for both classification and localization, as \emph{Conv2D} can extract spatial features while \emph{FC} and \emph{Conv1D} can not.

 We also test localization accuracy when summing attention maps from different layers in Eq.~\ref{eq:attention_map}. As shown in Fig.~\ref{fig:vis_cub_attention_sup} and Fig.~\ref{fig:vis_ilsvrc_attention_sup}, attention maps $\{A_*^1,...,A_*^L\}$ from different layers are complementary, so we summarize them for full object extent localization. Ablation studies in Table~\ref{tab:cub_ablation_attention_layer} and Table~\ref{tab:ilsvrc_ablation_attention_layer} demonstrate that summing all attention maps($A_*^l$) achieves the best localization performances. 

\subsection{Visualization Results}
\begin{figure*}[t]
    \centering
    \includegraphics[width=1.\linewidth]{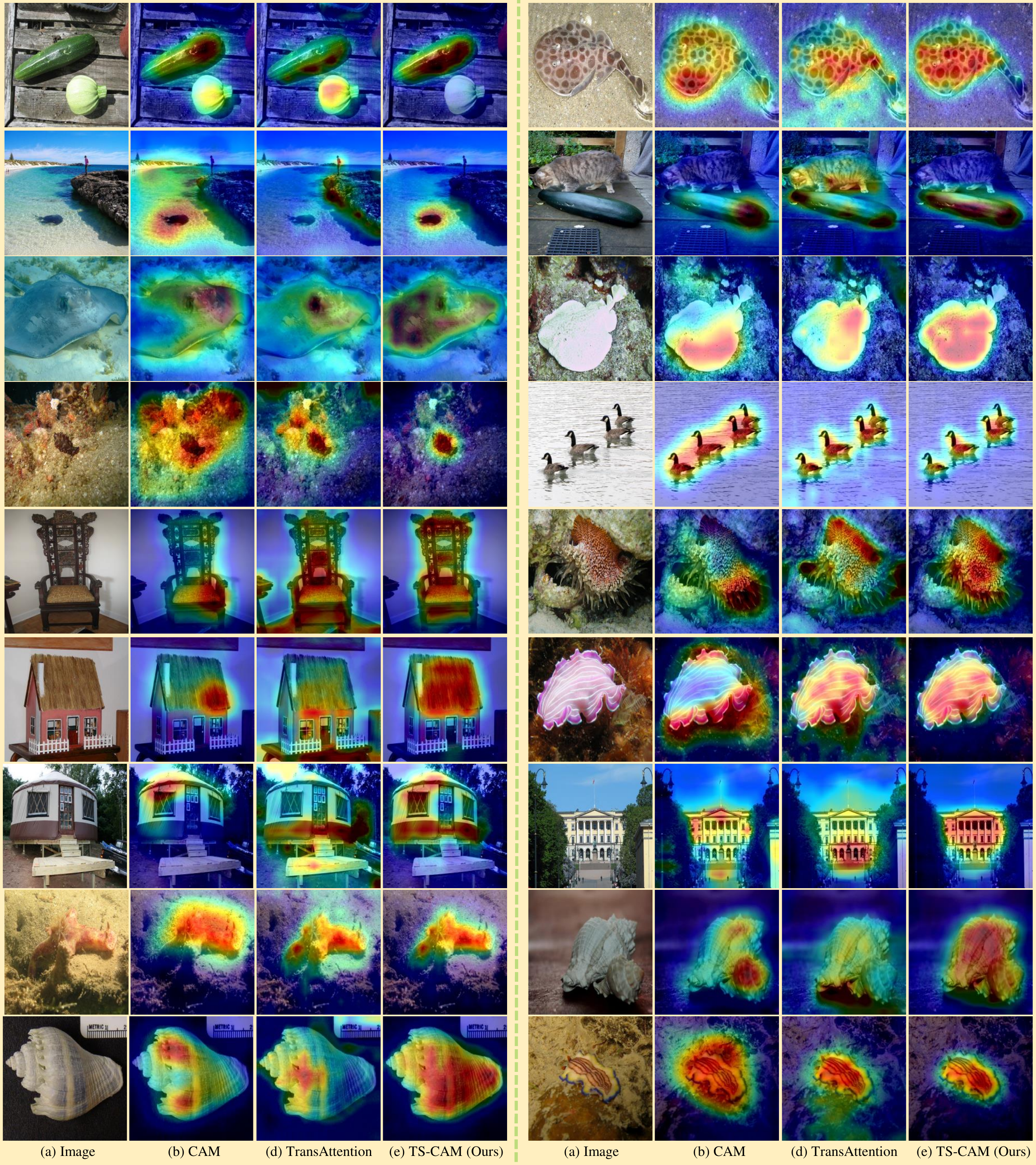}
    \caption{Visualization of localization maps on ILSVRC datasets. (a) Input Image. (b) Class Activation Map (CAM). (c) TransAttention:  Transformer-based Attention. (d) TS-CAM (ours). (Best Viewed in Color)}
    \label{fig:vis_ilsvrc_loc_sup}
\end{figure*}

\paragraph{Comparison of Localization Maps.}
Fig.~\ref{fig:vis_cub_loc_sup} and Fig.~\ref{fig:vis_ilsvrc_loc_sup} respectively show the additional localization results on CUB-2011-200 test set and ILSVRC validation set, where TS-CAM is able to extract the long-range feature dependency of visual transformer and localize object more accurately than CAM and \emph{TransAttention}. Specifically, the highlighted regions in the localization map of TS-CAM can outline the detailed contours of objects and preserve long-range feature dependency, while CAM~\cite{zhou2016learning} and \emph{TransAttention} fail. \emph{TransAttention} highlights incorrect foreground regions due to its lack of category semantics. 

\begin{figure*}[t]
    \centering
    \includegraphics[width=1.\linewidth]{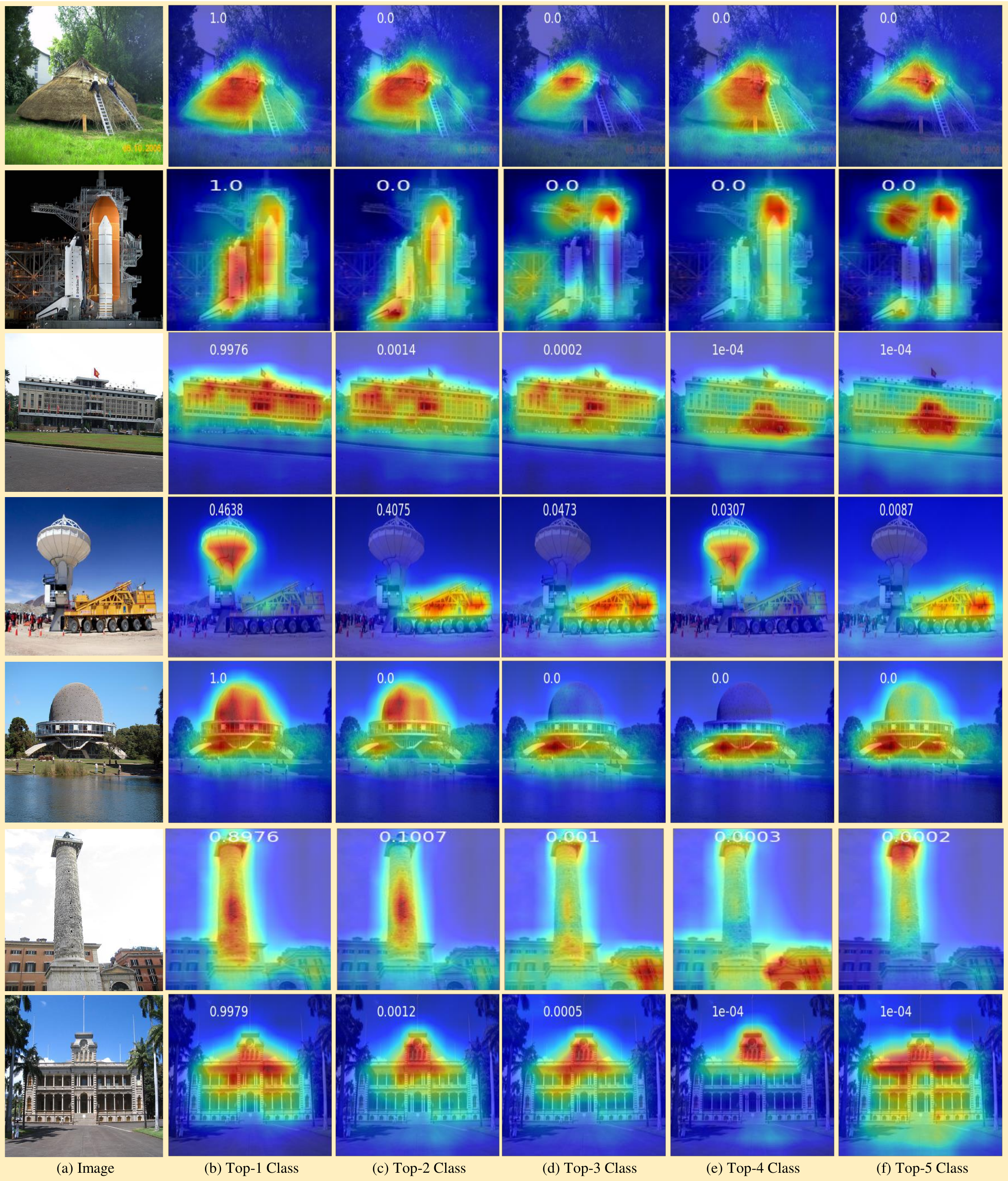}
    \caption{Visualization of localization maps for top 5 scored classes. The white numbers in left-top of the localization maps are the classification scores of each class, which are predicted by TS-CAM. (Best Viewed in Color)}
    \label{fig:vis_ilsvrc_top_loc_sup}
\end{figure*}

\paragraph{Top-5 Localization Maps.}
Fig.~\ref{fig:vis_ilsvrc_top_loc_sup} shows the localization maps of Top-5 scored classes. In these results, the localization maps of Top-1 scored class correctly highlight the object. We found that the most highlighted regions in the localization maps of the top-5 classes are different from each other for discriminative classification, while most of the maps highlight the similar regions since these 5 classes are semantic-similar.


\end{document}